\RequirePackage{amsthm}
\documentclass[sn-mathphys,Numbered]{sn-jnl}

\usepackage{graphicx}%
\usepackage{multirow}%
\usepackage{amsmath,amssymb,amsfonts}%
\usepackage{amsthm}%
\usepackage{mathrsfs}%
\usepackage[title]{appendix}%
\usepackage{xcolor}%
\usepackage{textcomp}%
\usepackage{manyfoot}%
\usepackage{booktabs}%
\usepackage{algorithm}%
\usepackage{algorithmicx}%
\usepackage{algpseudocode}%
\usepackage{listings}%

\usepackage{enumitem}
\usepackage{color}
\usepackage{soul}

\theoremstyle{thmstyleone}%
\theoremstyle{thmstyletwo}%

\theoremstyle{thmstylethree}%

\DeclareMathOperator*{\argmax}{argmax}

\usepackage{xcolor}      

\newcommand\myparagraph[1]{\vspace{\baselineskip}\noindent\textbf{#1}}

\newcommand\textcode[2]{\textcolor{#1}{\texttt{#2}}}
\newcounter{rqcounter}  
\newenvironment{rqlist}
  {\begin{list}
    {\arabic{rqcounter}}
    {\usecounter{rqcounter}
     \setlength{\leftmargin}{1.25cm}

    }
  }
{\end{list}}

\begin{document}

\title[Article Title]{Experiential Explanations for Reinforcement Learning}

\author*[1]{\fnm{Amal} \sur{Alabdulkarim}}\email{amal@gatech.edu}

\author[1]{\fnm{Madhuri} \sur{Singh}}
\author[1]{\fnm{Gennie} \sur{Mansi}}
\author[1]{\fnm{Kaely} \sur{Hall}}
\author[2,3]{\fnm{Upol} \sur{Ehsan}}
\author[1]{\fnm{Mark O.} \sur{Riedl}}

\affil[1]{\orgdiv{School of Interactive Computing}, \orgname{Georgia Institute of Technology}, \orgaddress{\city{Atlanta}, \state{Georgia}, \country{United States}}}

\affil[2]{\orgdiv{Berkman Klein Center}, \orgname{Harvard University}, \orgaddress{\city{Cambridge}, \state{Massachusetts}, \country{United States}}}
\affil[3]{\orgdiv{Khoury College of Computer Sciences}, \orgname{Northastern University}, \orgaddress{\city{Boston}, \state{Massachusetts}, \country{United States}}}
\abstract{

Reinforcement Learning (RL) systems can be complex and non-interpretable, making it challenging for non-AI experts to understand or intervene in their decisions. 
This is due in part to the sequential nature of RL in which actions are chosen because of their likelihood of obtaining future rewards.
However, RL agents discard the qualitative features of their training, making it difficult to recover user-understandable information for ``why'' an action is chosen. We propose a technique {\em Experiential Explanations} to generate counterfactual explanations by training {\em influence predictors} along with the RL policy. Influence predictors are models that learn how different sources of reward affect the agent in different states, thus restoring information about how the policy reflects the environment.
Two human evaluation studies revealed that participants presented with Experiential Explanations were better able to correctly guess what an agent would do than those presented with other standard types of explanation.
Participants also found that Experiential Explanations are more understandable, satisfying, complete, useful, and accurate. 
Qualitative analysis provides information on the factors of Experiential Explanations that are most useful and the desired characteristics that participants seek from the explanations.\footnote{Data availability statement: The authors declare that the code and data supporting the findings of this study are available within the paper and its supplementary information files.} 
}

\maketitle

\section{Introduction}\label{sec1}
\begin{figure}
  \includegraphics[width=\textwidth]{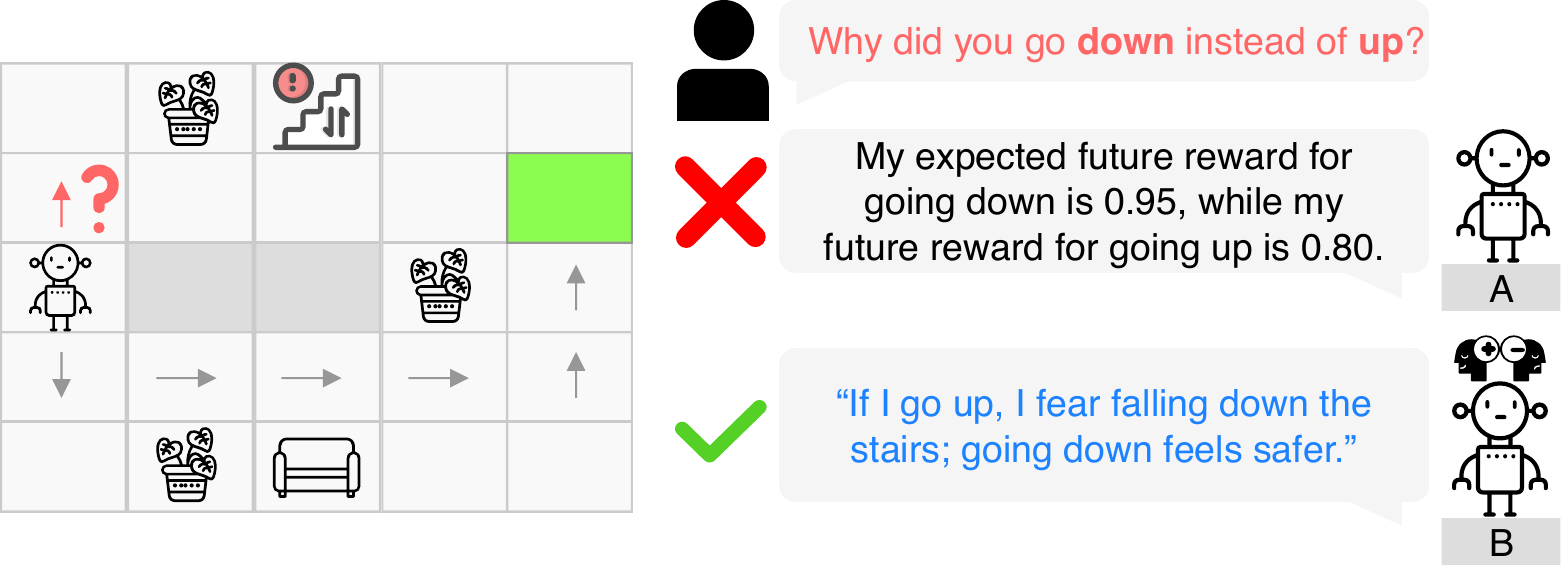}
  \caption{In an interaction between the user and the agent, the user expected the agent to go up, but the agent went down instead. The user asks the system for an explanation. The figure shows two types of explanations the user can receive. (A) is an explanation that can be derived directly from the agent's policy.
    (B) is an explanation that can be derived from the agent's policy with the assistance of additional models called negative and positive {\em influence predictors}. 
    }
  \label{fig:teaser}
\end{figure}
Reinforcement Learning (RL) techniques are becoming increasingly popular in critical domains such as robotics and autonomous vehicles. However,  applications such as autonomous cars and socially assistive robots in healthcare and home settings are expected to interact with non-AI experts. People often seek explanations for abnormal events or observations to improve their understanding of the agent and better predict and control its actions \cite{miller2019explanation}. This need for an explanation will likely arise when interacting with these systems, whether it is a car taking an unexpected turn or a robot performing a task unconventionally. Without an explanation, users can find it difficult to trust the agent's ability to act safely and reasonably ~\cite{zelvelder2021assessing}, especially if they have limited artificial intelligence expertise. 
Even in the case where the agent is operating optimally and without failure, the agent's optimal behavior may not match the user's expectations, resulting in confusion and lack of trust.
Therefore, RL systems need to provide good explanations without compromising their performance. 
To address this need, {\em explainable RL} has become a rapidly growing area of research with many open challenges \cite{anjomshoae_explainable_2019,alharin2020reinforcement,wells_explainable_2021,heuillet_explainability_2021,puiutta_explainable_2020,milani_survey_2022,chakraborti_emerging_2020}.

But what makes a good explanation for end-users who cannot change the underlying model? 
How do we generate explanations if we cannot modify the underlying RL model or degrade its performance?
In RL, an agent learns dynamically  to maximize its reward through a series of experiences interacting with the environment~\cite{sutton2018reinforcement}. 
Through these interactions, the agent builds a utility model, $V(s,a)$, which estimates the future reward that can be expected to be achieved if a particular action $a$ is performed from a particular state $s$.

While an RL agent learns from its experiences during training, those experiences are inaccessible after their training is over~\cite{alharin2020reinforcement}. 
This is because $V(s,a)$ summarizes future experiences as a single, real-valued number; this is all that is needed to execute a policy $\pi = \mathrm{argmax}_a V(s,a)$.
For example, consider a robot that receives a negative reward for being close to the stairs and thus learns that states along an alternative route have higher utility. 
When it comes time to figure out {\em why} the agent executed one action trajectory over another, the policy is devoid of any information from which to construct an explanation beyond the fact that some actions have lower expected utility than others.

Explanations need not only address agents' failures. The user's need for explanations can also arise when the agent makes an unexpected decision. Requests for explanations thus often manifest themselves as ``why-not'' questions referencing a counterfactual: ``why did not the agent make a different decision?''
Explanations in sequential decision-making environments can help users update their mental models of the agent or identify how to change the environment so that the agent performs as expected. 

We propose a technique, {\bf Experiential Explanations}, in which deep RL agents generate explanations in response to on-demand, local, counterfactuals proposed by users. These explanations qualitatively contrast the agent's intended action trajectory with a trajectory proposed by the user and specifically link agent behavior to environmental contexts.
The agent maintains a set of models---called {\bf influence predictors}---of how different sparse reward states influence the agent's state-action utilities and, thus, its policy. 
Influence predictors are trained alongside an RL policy with a specific focus on retaining details about the training experience.
The influence models are ``outside the black box'', but provide information with which to interpret the agent's black-box policy in a human-understandable fashion. 

Consider the illustrative example in Figure \ref{fig:teaser} where the user observes the agent go down around the wall. 
The user expects the agent to go up, but the agent goes down instead.
The user asks why the agent did not choose up, which appears to lead to a shorter route to the goal.
One possible and correct explanation would be that the agent's estimated expected utility for the down action is higher than that for the up action. 
However, this is not information that a user can easily act upon. 
The alternative, Experiential Explanation, 
states that up will pass through a region that is in proximity to dangerous locations.
The user can update their understanding that the agent prefers to avoid stairs.
The user can also understand how to take an explicit action to change the environment, such as blocking stairs---an action a non-expert can take to influence agent behavior when they do not have the ability to directly change the agent.
Our technique focuses on post-hoc, real-time explanations of agents operating in situ, instead of pre-explanation of agent plans~\cite{chakraborti2021emerging}, although the two are closely related.

We evaluate our Experiential Explanations technique with studies with human participants. We show that explanations generated by our technique allow users to better understand and predict agent actions and are found to be more useful. We additionally perform a qualitative analysis of participant responses to elicit {\em how} users use the information in the explanations in our technique and baseline alternatives to reason about the agent's behavior.
This not only provides evidence for Experimental Explanations, but provides a blueprint to understand the human factors of other explanation techniques.

\section{Background and Related work}
Reinforcement learning is an approach to learning in which an agent attempts to maximize its reward through feedback from a trial-and-error process. 
RL is suitable for Markov Decision Processes, which can be represented as a tuple $M =\langle S, A, T, R, \gamma \rangle$ where $S$ is the set of possible world states, $A$ is the set of possible actions, $T$ is the state transition function, which determines the probability of the following state $P(S)$ as a function of the current state and action.  $T : S \times A \rightarrow P(S)$, $R$ is the reward function $R : S \times A \rightarrow \mathbb{R}$, and $\gamma$ is a discount factor $0 \leq \gamma \leq 1$. RL first learns a policy $\pi : S \rightarrow A$, which defines which actions should be taken in each state. Deep RL uses deep neural networks to estimate the expected future utility such that $\pi(s) =\argmax_a V_\theta(s,a)$ where $\theta$ are the model parameters.

As deep reinforcement learning is increasingly used in sensitive areas such as healthcare and robotics, Explainable RL (XRL) is becoming a crucial research field. 
However, the reinforcement learning paradigm differs substantially from other machine learning paradigms, such as supervised learning, in that they solve sequential decision-making problems where each decision an agent makes is not an end to itself but establishes conditions for future actions; understanding any one decision may be dependent on understanding prior decisions as well as future potential decisions.
For an overview of XRL, please refer to \citeauthor{milani_survey_2022} \cite{milani_survey_2022}. We highlight the most relevant work on XRL to our approach.

Explanation via Reward Decomposition breaks down rewards into a sum of semantically meaningful reward categories, allowing actions to be evaluated based on the trade-offs between these categories \cite{juozapaitis2019explainable}.
Many XRL techniques generate explanations by decomposing utilities in the agent's network and showing information about the components of the utility score for an agent's state~\cite{das2020leveraging,anderson2019explaining, septon2023integrating}.
Our technique is related to decompositions in that we learn the influence of different sources of reward on utilities.
However, we differ in that we approximate the decomposition using external models, avoiding the trade-off between using a semi-interpretable system or a better black-box system.

Explanations are inherently contrastive, as they are framed in the context of particular counterfactual comparisons. Rather than merely questioning why event P took place, individuals ask why event P occurred rather than event Q \cite{miller2019explanation}. Contrastive approaches such as
\citeauthor{van2018contrastive} \cite{van2018contrastive} leverage an interpretable model and a learned transition model. \citeauthor{madumal2020explainable} \cite{madumal2020explainable} explain local actions with a causal structural model as an action influence graph to encode the causal relations between variables. 
\citeauthor{sreedharan2022bridging} \cite{sreedharan2022bridging} generate contrastive explanations using a partial symbolic model approximation, including knowledge about concepts, preconditions, and action costs. These methods provided answers to user's local why not questions. While our approach still produces counterfactual explanations, we minimize the use of predefined symbolic or {\em a priori} environment knowledge. Other work looked at utilizing counterfactual explanations for other explanation goals. 
For example, \citeauthor{frost2022explaining} \cite{frost2022explaining} trained an exploration policy to generate test time trajectories to explain how the agent behaves in unseen states after training. Explanation trajectories help users understand how the agent would perform under new conditions. While this method gives users a global understanding of the agent, our method focuses on providing local counterfactual explanations of an agent's action. 
\citeauthor{olson2021counterfactual} \cite{olson2021counterfactual} explains the local decisions of an agent by demonstrating a counterfactual state in which the agent takes a different action to illustrate the minimal alteration necessary for a different result. \citeauthor{huber2023ganterfactual}\cite{huber2023ganterfactual} furthers this concept by creating counterfactual states with an adversarial learning model. Our technique, however, concentrates on the qualitative distinctions between trajectories and not on the states.

\textit{Rationale generation}, introduced by \citeauthor{ehsan2019automated}\ \cite{ehsan2019automated}, is an approach to explanation generation for sequential environments in which explanations approximate how humans would explain what they would do in the same situation as the agent. 
Rationale generation does not ``open the black box'' but looks at the agent's state and action and applies a model trained from human explanations.
Rationales do not guarantee faithfulness, though studies show they can still provide actionable insights.
Like rationale generation, our models train alongside the agent.
However, our technique also achieve a high degree of faithfulness.
State2Explanation (S2E), introduced in \citeauthor{Das2023State}\cite{Das2023State}, demonstrates generating more faithful explanations by using them to inform the agent's reward shaping and requires the derivation of the concepts from mathematical representations and expert domain knowledge. By aligning state-action pairs with high-level concepts, S2E uses natural language templates to provide actionable insights during training and deployment. In our approach, we avoid intervening with the learning of the agent while preserving a high level of faithfulness by confining the learning only from the agent's transitions and trying to minimize the reliance on domain knowledge to only the names and effects of reward components.

\section{Experiential Explanations}

{\em Experiential Explanations} are contrastive explanations generated for local decisions with minimal reliance on structured inputs and without imposing limitations on the agent or the RL algorithm. Our explanation technique uses additional models called {\em influence predictors} that learn how the sources of the sparse rewards affect the agent's utility predictions. 
The agent's learned policy learns the utilities with respect to all rewards. 
Influence predictors can tell us how strongly or weakly any source of reward (positive or negative) is impacting states in the state space that the agent believes it will pass through.
The influence predictors provide the additional context that explains the agent's choices through finer-grained utility detail. As in Figure~\ref{fig:teaser}, we see the explanation referencing the agent's relationship to the environment in terms of negative and positive elements.

The main components of Experiential Explanations are the trained influence predictors and the two modules of explanation generation. The first module starts with the user proposing an alternative action and asking why the agent did something else; seeking a counterfactual explanation. This triggers simulation of the trajectories of both actions. For example, in Figure~\ref{fig:teaser} the agent's action is down, but the user asks about going up. The generated trajectories account for those actions and continue until the end of the respective episodes. In the second module, we compare the states along a specified horizon of the simulated trajectories using influence predictors.

In Section \ref{sec:influence_predictors} we dive into the deeper details of the influence predictors. In Section \ref{sec:explanation_generation} we demonstrate how they can be used in the explanation generation modules to generate explanations. 

\subsection{Influence Predictors}
\label{sec:influence_predictors}
{\em {Influence predictors}} reconstruct the effects of rewards received on the utility learned by the agent during training. These models are trained alongside the agent to predict the strength of the influence of different sources of reward on the agent by learning a value function that outputs the expected utility from each source of reward: $U_{c}(s, a)$ where each $c \in C$ is a distinct source (or class) of reward. For example, a terminal goal state might be a source of positive reward, stairs might be a source of negative reward, and other dangerous objects may be other sources of negative reward. In a more complex environment, multiple sources of positive and negative rewards can be helpful in describing the agent's plan. For instance, a robot attempting to make a cup of coffee would receive positive rewards for obtaining a clean mug, hot milk, and coffee, then combining the milk and coffee in the mug and stirring them. Negative influences could include getting a dirty cup or adding salt.

Each source of reward has its own influence predictor model. Typically, one class is associated with a goal state (or states required for task completion) and produces positive rewards. 
Sometimes an intermediate positive reward is given when it is known in advance that the task requires particular actions or states to be traversed, as in the case of reward shaping \cite{ng1999policy}.
Other classes are associated with states that are {\em a priori} known to be detrimental to the task and produce negative rewards.

In principle, Influence Predictors can be used with any agent architecture and optimization algorithm as long as we can observe its transitions during training.
Influence predictors are trained similarly to an RL agent (e.g., standard Deep $Q$-networks (DQN)~\cite{mnih2015human}, or A2C \cite{mnih2016asynchronous}) but with two differences.
First, influence predictors are {\em not} used to guide the agent's exploration during training or inference. They rely on the main policy's training loop to generate transitions. Second, each influence predictor model can receive only rewards from one class of rewards. Influence predictors are trained using loss based on a modified Bellman equation:

\begin{equation}
U_{c}(s_t,a_t) = |r^{c}_{t+1}|+\gamma \max_{a'}U_{c}(s_{t+1}, a') 
\end{equation}
where $r^{c}_{t+1}$ is the reward associated with $c$ delivered to the agent upon transition to state $s_{t+1}$. The absolute value re-interprets the reward as an influence, since the influence predictor is now predicting the utility of attaining the reward, even if it is negative. As a consequence of ever receiving only the absolute value of a class of rewards, $U_{c}(\cdot)$ is an estimate of the strength of $c$ in $s$ and $a$.

The training process for influence predictors is shown in Algorithm ~\ref{fig:training-proc}. The agent, controlled by its base policy learning algorithm, interacts with the environment, executing actions, and receiving state observations. State transitions, actions, and rewards are recorded and used to populate separate buffers for each influence predictor. 
For model updates, we sample the transitions from each designated buffer separately and backpropagate loss in each influence predictor. 

\begin{algorithm}[H]
\scriptsize
\caption{ Influence predictor and agent training}
\label{alg:cap}
\begin{algorithmic}[0]
\For{each $c \in C$ }  \Comment{$C$ is the reward sources}
\State  Initialize buffer memory $D_c$ to capacity $N$
\State  Initialize action-value function $U_c$ with random weights
\EndFor
\For{episode = 1, M} 
    \State $transitions_{T} \gets agent.transitions(T)$  \Comment{The agent's transition data from its buffer} 
    \For{t=0, T} 
        \For{each $c \in C$} 
            \State $transition_t[c] \gets transitions_{T}[t]$ 
            \If{$reward_t \in c$} 
                \State $transition_t[c][reward] \gets \lvert transitions_{T}[t][reward]$$ \rvert$ 
            \Else
                \State $transition_t[c][reward] \gets 0$
            \EndIf
            \State $(transition_t[c]) \to D_c$
    \EndFor
\EndFor
\For{each $c \in C$} 
    \State $U_c \gets update(D_{c})$
\EndFor    
\State $agent \gets update(transitions_{T})$
\EndFor
\end{algorithmic}
\label{fig:training-proc}
\end{algorithm}

\begin{figure}[!t]
    \centering
    \includegraphics[width=\columnwidth]{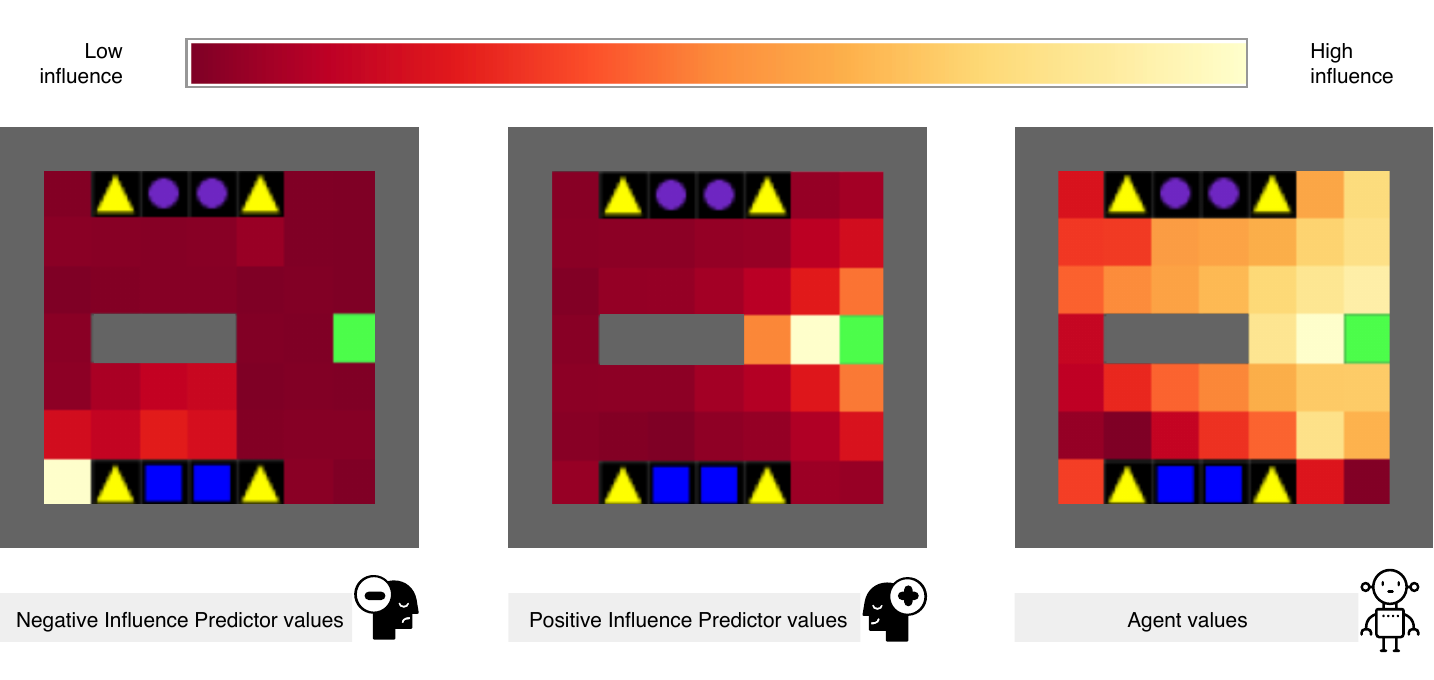}
    \caption{This figure shows an example of the agent's learned maximum learned utility value at each state and the two influence predictors values visualized on heatmaps. The blue squares are the source of negative rewards, whereas the green square is the source of positive rewards. The purple and yellow shapes have no effect.
    Intuitively, one can see that the agent's policy (right) is closely associated with the positive influence values and the {\em opposite} of the negative influence values.
    }
    \label{fig:qvalues-comparison}
\end{figure}
To illustrate, Figure~\ref{fig:qvalues-comparison} shows an agent operating in an environment with two reward classes: a positive reward class for the terminal goal, and a negative reward class for the blue squares---hence, two influence predictors. The influence map on the left was generated by querying the negative influence predictor for each state and maxing across all possible actions. 
The negative influence trends are stronger closer to the blue squares. 
The influence map in the middle is generated from the positive influence predictor. 
The map on the right is the agent's main policy model, trained on all reward classes as normal. Even though the visualization aggregates action utilities, one can see that the agent will prefer a trajectory in the upper part of the map.

\subsection{Explanation Generation}
\label{sec:explanation_generation}
At any time during the execution of the policy, the user can request a counterfactual explanation by proposing an action $(a_\mathrm{user})$ rather than the one the agent chose $(a_\mathrm{agent})$. Given the action chosen by the agent and the alternative action proposed by the user, the explanation generation process proceeds in three phases. 

First, we generate and record the trajectory $\tau_\mathrm{agent}$ the agent will take from this point forward.\footnote{We assume access to a simulator or a world model (e.g., 
\cite{hafner2020dreamerv2}) with sufficient fidelity to predict action outcomes, though a world-model based learner was not used to create a minimalistic experiment.}
Second, we generate a counterfactual trajectory $\tau_\mathrm{user}$ as if the agent had executed the user's proposed alternative action. To generate $\tau_\mathrm{user}$, we force the agent to follow the user's suggested alternative action (or sequence of actions in the case that the user wants to specify a partial or complete counterfactual trajectory) and complete the trajectory using the same planning approach. 

Finally, we use influence predictors to compare the two trajectories. Recall, in Section \ref{sec:influence_predictors}, we have an influence predictor $IP_c$ per reward source $c \in C$. We have access to all the influence predictor values in both trajectories at this stage, which gives us control over the types of explanation we can generate depending on our use case. 

We can produce local, detailed, global, or aggregated explanations in natural language.
To do so we require one modification to the typical Markov Decision Process formulation:
we use an {\em enhanced} reward function, 
\begin{equation}
R: S \times A \rightarrow \mathbb{R} \times \mathcal{L}^*
\end{equation}
where $\mathcal{L}$ is a natural language vocabulary and $\mathcal{L}^*$ is the set of all possible natural language strings.
The enchanced reward function produces both a real valued number and a natural language string that describes the circumstances under which the reward is given.
For example, for the scenario in Figure~\ref{fig:teaser}, the enhanced reward function might produce the value $-1$ and ``fall down stairs'' when the agent encounters the location in the top center of the environment.

These textual descriptions of reward circumstances can be embedded in the explanation templates to automatically add stance or other details about the reward influences to the explanations. 
In this work, we explore two variants of the explanations: an {\em aggregated explanation} and a detailed {\em local explanation}.

\textbf{Aggregated Explanation.} 
An aggregated explanation presents an overview of the effect of each reward class on the agent throughout a complete trajectory. To generate these explanations, we calculate the mean value of each influence predictor $IP_{c}$ for all states 
$s \in \tau_\mathrm{agent}$ 
and compare this mean with the other trajectories. 
\begin{equation}
\argmax_{a \in \{a_\mathrm{user}, a_\mathrm{agent}\}} (\frac{\sum_s{\max(IP_{c}(\tau_\mathrm{a}[s]))}}{length(\tau_\mathrm{a})}), \forall c \in C.
\end{equation}
The resulting explanation assigns each influence predictor to the trajectory in which they were the most dominant. We then take these values and translate them into natural language (English) explanations. If the values generated by $IP_{c}$ were similar for both trajectories, then they can be excluded from the explanation, as they do not provide valuable, contrastive information. For example, in the case of two influence predictors in a simple environment, if we assume that we only had a difference in the negative influence predictor, the explanation would be as follows: \textit{``If I go up, I fear falling down the stairs; going down feels safer''}. This case is demonstrated in Figure \ref{fig:teaser}. Suppose that there is a difference in an additional $IP_{c}$ mean. In that case, we add it to the explanation by appending its relevant sentence to explain all influence predictors if we observe a significant difference in their mean values.

\textbf{Local Explanation.} In more complex environments where there are many influences, the explanations can shed light on the agent's near-horizon and concentrate more on exploring local values. These local explanations zoom in on the most influential factors related to actual and counterfactual actions over their corresponding {\em segments}, which are denoted by $t_\mathrm{agent} \subseteq \tau_\mathrm{agent}$. Each segment starts at the beginning of the designated action and includes a pre-determinted number of additional steps (we used five steps) after the end of the action to capture some of the longer-term effects.

To generate a local explanation, we first identify the most significant factors for each local segment by calculating the maximum influences of each state $s$ during a segment:
\begin{equation}
\label{equation:argmax_ip_sig}
\{\argmax_{c \in \{C\}} (\max{(IP_{c}(s))}) | \forall a \in \{a_\mathrm{user}, a_\mathrm{agent}\}, \forall s \in t_\mathrm{agent} \}
\end{equation}
From equation \ref{equation:argmax_ip_sig} we get two sets of influence predictors that had the maximum value at any state within the segments $t_{agent}$ and $t_{user}$. 
If there is no difference between the segments of $a_\mathrm{agent}$ and $a_\mathrm{user}$ with respect to their sets of maximum influences $IP_{c}$, we resolve to compare the mean values of the influence predictors during the same period and obtain the three influences with the highest means in their segments. First, we get the mean of every influence predictor in each segment:
\begin{equation}
mean(IP_{c}, a) = \frac{\sum_{s}{\max{(IP_{c}(t_\mathrm{agent}[s]))}}}{||t_\mathrm{agent}||}, \forall c \in C, \forall a \in \{a_\mathrm{user}, a_\mathrm{agent}\}
\end{equation}
Then, we obtain the top three influences for each segment $t_\mathrm{agent}$ and $t_\mathrm{user}$, ordered by $mean(IP_c)$.

We provide this explanation even if there is no difference in the top influences, which can happen if the alternative action considered is too similar to the actual action taken. To generate English explanations, we use the list of the top influence predictors obtained through the maximum- or highest-mean-method. And then embed them into the templates. An example of a generated explanation can be: \textit{``If I got the nearby cup, I would still have to clean it before making my coffee. But if I get a cup from the cupboard, I will not need to clean it and I can immediately make coffee.''}. Where ``cleaning'' and ``making coffee'' are terms derived from the enhanced reward sources, as described above.

The explanation strategies mentioned above provide information about how external environmental rewards affect the choices made by the agent, even if they are not directly linked to the agent's actions. These explanations are in line with the definition of first-order explanations \cite{dazeley2021levels}, as they reveal the underlying inclinations of the agent towards the environment and other factors that influence its decisions.

\section{Evaluation overview}

Our evaluations focus on exploring how beneficial Experiential Explanations are to participants in understanding RL agents and how they use these explanations to reason about their own decisions through quantitative and qualitative analyzes. We conducted two studies to evaluate our explanations with human participants in {\em simple} and {\em complex} environments. 
Our simple environment is a grid world built on Minigrid~\cite{minigrid-2018} environment.
Our complex environment is the Crafter game~\cite{hafner2021crafter}, a 2D version of the popular commercial game, Minecraft.

Our evaluations address the following research questions in both settings.

\begin{rqlist}
\item Do explanations derived from influence predictors accurately represent the actual values of the agent? 
\item How do Experiential Explanations improve users' ability to predict the agent's actions compared to baselines? 
\item How does user satisfaction with Experiential Explanations compare to baseline explanations? 
\item How do users who got Experiential Explanations compare in their reasoning with those who received baseline explanations? 
\end{rqlist}
In both simple and complex environments, we first address {\bf RQ1}, which focuses on examining the faithfulness of the explanation, a key consideration of explanation generation, which measures the extent to which the explanations are faithful to the underlying system \cite{zhou2021evaluating}.
Influence predictors sit ``outside of the black-box'', whereas the main policy model that drives the agent is ``inside''.
However, influence predictors are trained alongside the main policy model, and thus, if they learned correctly, should be able to estimate the agent's actions with a high degree of accuracy.
We don't expect them to replicate the agent's actions as they are learning influences instead of generating actions.
To assess the faithfulness of influence predictors, we evaluated the accuracy of the combined decision of influence predictors in predicting the agents' actions and how closely the combined influence predictor's values estimate the agent's. 

We then designed and conducted a our human-participant study following the framework proposed by \citeauthor{Hoffman2018MetricsFE} \cite{Hoffman2018MetricsFE}, evaluating our explanations for actionability, satisfaction, and utilization. We examine the actionability of the given explanations through a prediction and ordering tasks of the agent's actions and sub-goals ({\bf RQ2}). Then we examine the participants' perception of the explanations and their satisfaction ratings ({\bf RQ3}).  Lastly, both the prediction tasks and the satisfaction survey are followed by reasoning questions that we qualitatively analyze to explore the use of explanation elements in reasoning for the different groups ({\bf RQ4}). 

The first experiment focused on understanding the overall utility of the explanations in understanding and predicting the agent's actions. This experiment is set in a MiniGrid environment, where we examine the performance of the aggregated version of the explanations in a sparse-reward setting with two reward classes. 

In the second study, we evaluate the usefulness and actionability of explanations in a more complex environment. We set our experiment in the Crafter environment to explore the effectiveness of the local explanations generated by our Experiential Explanations method in a more complex environment with sequential multistep goals, with more granular reward classes. Moreover, we want to see if the local Experiential Explanations can convey these complex values to the participants coherently.

In the next sections, we provide a comprehensive overview of the two experiments we designed and conducted to evaluate our explanation technique and address our research questions.

\section{Experiment 1: MiniGrid Environment}

\begin{figure}
\centering
\includegraphics[width=\columnwidth]{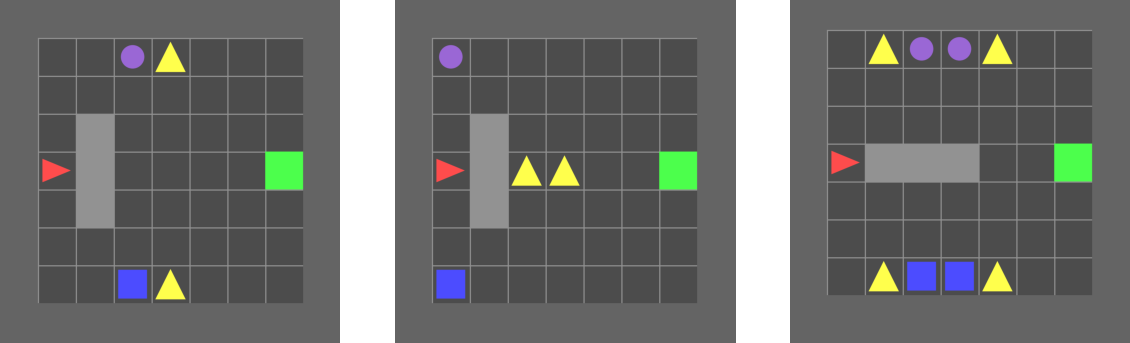}
\caption{The three episodes from the human study, in order presented to participants.}
\label{fig:cases}
\end{figure}
In this study, we evaluate our explanations in an environment with two simple reward sources and simplified grid-navigation settings, with only one possible counterfactual action at the decision point (see Figure~\ref{fig:cases}). To answer our research questions we first automatically validated the faithfulness of the explanations 
(RQ1) and then we evaluated our explanation method with quantitative and qualitative analyses through a human evaluation study (RQ2, RQ3 and RQ4). We built our experimental domain on MiniGrid~\cite{minigrid-2018}, a popular minimalistic grid world environment that offers a variety of tasks to test RL algorithms.

\subsection{Model Setup and Environment Details}   
 \textbf{Environment customization.} We require an environment with multiple reward classes---positive rewards as well as one or more sources of negative rewards---thus we created a custom task where the agent is placed in a room with a goal (a green square) and other objects represented by colored shapes that we could assign to neutral or negative reward values when approached. 
The environment provides sparse rewards; a positive reward of $(1- \frac{steps}{\text{max allowed steps}})$ for reaching a green goal tile (the minigrid default) and $-1$ for stepping into a dangerous object.
Both the goal and stepping on the dangerous object terminate the episode. 
All other states and actions produce zero reward.

\myparagraph{Agent training.} Our base agent policy model is the advantage actor-critic (A2C)~\cite{willems-2018} 
optimized using the exponential policy optimization algorithm (PPO) \cite{schulman2017proximal}. 
A state observation is an image of the grid constrained by a limited field of view. 
We include more technical details about the agent's architecture and training in Appendix \ref{appendix:agent}. 

\myparagraph{Influence Predictors design and training.} We are using DQNs as our influence predictors (as described in \cite{mnih2015human}). Specifically, we have two predictors, one for negative rewards and one for positive/goal rewards. We provide more details on the training and architecture of the Influence Predictors in Appendix \ref{appendix:influence_predictor}. 

\myparagraph{Explanation generation.} We use the aggregated explanations described in Section \ref{sec:explanation_generation}. These explanations contrast the aggregated difference in the means between the agent's paths, the one associated with the proposed counterfactual, and the one the agent did. Below are some examples of these explanations:
\begin{enumerate}
\item \textbf{Negative rewards heavily influence the counterfactual trajectory.} \\\textbf{Template:}
\textit{``If the agent $[a_{user}]$, it will pass through regions influenced by the [negative influence]; going $[a_{agent}]$ feels safer''}.
\\\textbf{Explanation Example:}\textit{``If the agent goes down, it will pass through regions influenced by the dangerous blue obstacles; going up feels safer''}.
\item \textbf{Goal rewards have a lower influence on the counterfactual trajectory.} 
\\\textbf{Template:}
\textit{``If the agent $[a_{user}]$, it will pass through regions less influenced by the [positive influence]; going $[a_{agent}]$ feels better''}.
\\\textbf{Explanation Example:}
\textit{``If the agent goes right, it will pass through regions less influenced by the green goal; going left is better''}.
\end{enumerate}

\subsection{Baselines}
In this experiment, we focus on similar baseline explanations that the agent can generate. We compared our method with the following baselines: 
\begin{itemize}
\item \textbf{Heatmap explanation:} Showing the agent's learned maximum values for each state. We generate the explanations for this baseline by getting the maximum agent value in each possible state on the map. This is similar to the agent values map in Figure \ref{fig:qvalues-comparison}.
\item \textbf{Q-Value explanation:} Presenting the average expected reward for both paths.
These explanations utilize the simulator from our explanation generation pipeline, but not the influence predictors. This baseline is a comparison of the mean agent's values between the paths. An example of such an explanation is \textit{The Q average if the agent goes up  9.84, while the Q average if the agent goes down is 9.66.} 
\item \textbf{No explanation:} The users only saw the correct answer without further explanation.
\end{itemize}

\subsection{Faithfulness Evaluation Methodology}
\label{faith_method}
To evaluate the reliability of our explanation method, we resort to addressing the faithfulness ({\bf RQ1}), we combine the predictions of the influence models using two methods. 
The first is to combine the values of the influence predictors and using their cumulative value to estimate the agent's action $action = \argmax_{a}(\sum_{i}P_{i}(s,a) - \sum_{j}N_{j}(s,a))$,
where $P_{i}$ is a positive influence predictor and $N_{j}$ is a negative influence predictor.
The second method is to train a classifier on the predicted positive and negative influence values as features and the agent's actions as the ground truth. We used a support vector machine (SVM) classifier with a radial basis function (RBF) kernel.
We also hypothesized that in the sparse reward setting, it is more meaningful to look at the influences around the emitting rewards rather than further away from them, as it would likely be more influential to the agent's decision in those areas rather than further away and therefore, the values would likely be better at predicting the agent's actions when those influence signals are high enough to impact the agent. To do this, we filter our dataset, by influence values at a set of different fixed intervals and examine the prediction accuracy at each threshold, we start at $0.100$ and we keep increasing by $+0.1$ until it is no longer meaningful to do so.
The faithfulness methodology does not make use of the baselines since it is the comparison of the aggregation of information from influence predictors to the main policy model.

\subsection{User Study Methodology}
We designed our human-participant evaluation following the framework proposed by \citeauthor{Hoffman2018MetricsFE} \cite{Hoffman2018MetricsFE}, evaluating our system for understandability, performance, and user satisfaction. We conducted our online evaluation between subjects using Prolific\footnote{http://prolific.co} with 81 participants
aged 21-71 years old (M=37.41, SD=11.03); 39.5\% of the participants identified as women. The average study completion time was approximately 10 minutes, and we compensated each participant with \$3.75. 
The study consists of two parts: answering a series of prediction tasks (RQ2, RQ4) and a satisfaction survey (RQ3, RQ4). The explanation group selection were fully randomized and balanced using Qualtrics\footnote{https://www.qualtrics.com/}.

The study involved participants observing an agent (represented as a red triangle) navigate towards a goal (green square) across three episodes, with certain shapes yielding negative rewards. Initially, participants made uninformed predictions about the agent's movements, which were manipulated to ensure incorrectness. This established the conditions for which the participant must now learn to correct their understanding of the agent's interactions with the environment. This setup was used to teach the participants to adjust their understanding based on consistent negative reward assignments in subsequent episodes. The last two episodes were considered for the prediction task. The expectation was that the participants in the experimental condition would demonstrate improved prediction accuracy over those in the baseline conditions. In addition, a satisfaction survey assessed participants' perceptions of the explanations provided for the agent's behavior. Each of the questions in the prediction task we followed by a text entry question to capture the participants rationales for making these predictions. In addition, another text entry question was added to get an overall impression of the helpfulness of these explanations. These are the responses that we analyze in our qualitative section.  We include screenshots and a detailed description of the study in the Appendix  \ref{appendix:eval-survey1}.

\begin{table}[t]
\centering
\begin{tabular}{lccc}
\toprule
           & $2^{nd}$ Correct  & $3^{rd}$ Correct  & Both Correct    \\
\midrule
Experiential Exp.    &       \textbf{90.48\%} &        \textbf{95.24\%} &      \textbf{85.71\%} \\
Heatmap Exp. &       78.95\% &        78.95\% &      63.16\% \\
Q-value Exp. &       80.95\% &        80.95\% &      66.67\% \\
No Explanation      &       75.00\% &        70.00\% &      55.00\% \\
\bottomrule
\end{tabular}
    \caption{The percentages of correct answers for first prediction task and second prediction task and the combined percentages. }
    \label{tab:correcntess-percentage}
\end{table}

\subsection{Quantitative Results}

\myparagraph{RQ1: Faithfulness.}
For measuring faithfulness we look at the accuracy of influence models in predicting agent actions, we found that the accuracy of the prediction of the direct aggregation is $72\%$ and the accuracy of the classifier $76\%$. 
To examine the alignment of the influence predictors with the agent at states with higher influence, we gradually examine the alignment between the highest action as predicted by the influence predictors and the agent's actions. At a threshold of $0.100$, the aggregate of influence predictors agreed with the main policy $83\%$, covering at least $20\%$ of the environment. 
At higher thresholds, agreement increases above $90\%$, indicating that when decisions are the most important, faithfulness is very high.
Looking at the positive influence predictor alone, at a threshold of $0.200$ and higher, it alone can predict the agent with a precision of $100\%$. 
This is to be expected; influence predictors cannot dictate what states and actions get explored---it is controlled by the main policy and the main training algorithm---so states that are less critical to positive reward attainment are under-sampled and prone to more error in both the main policy and the influence predictors.
In contrast, more important decision points are sampled more often, so the influence predictors get more training on those states and actions.

\myparagraph{RQ2: Prediction Correctness.}
We measure whether the participant can correctly predict the agent trajectory after interacting with the second and third episodes (Figure \ref{fig:cases}). 

Table \ref{tab:correcntess-percentage} shows the percentage of correct answers for each type of explanation after the second and third episodes (the first was a control), as well as the combined average.
Participants provided with Experiential Explanations had the highest success rate. 
Experiential Explanations were significantly higher than the no-explanation condition, validated with a logistic regression $(p < 0.05, df = 3, \chi^2 =  4.404)$.
Experiential Explanations led to a 9.5+\% improvement over other baselines, but were not statistically significant at the same threshold.

\myparagraph{RQ3: Satisfaction Survey.}
For each type of explanation, we analyzed the ratings of the participants for the Likert scale questions on the satisfaction survey. We include a visualization of the results in Figure \ref{fig:survey} in the Appendix. Experiential Explanations had the highest average scores on all dimensions, except trust. A one-way analysis of variance (ANOVA) followed by Tukey's honestly significant difference test revealed that the Experiential Explanations  $(M=3.62, SD=1.32)$ were preferred over the Heatmaps  $(M=2.42, SD=1.35)$  in completeness ($F(3, 81)= 3.032, p < 0.05$). 
Experiential Explanations  $(M=3.71, SD=1.0)$  were preferred over the Heatmaps  $(M=2.63, SD=1.38)$  in satisfaction($F(3, 81)=  3.106, p < 0.05$). 
Experiential Explanations $(M=4.38, SD=0.59)$ were preferred over Q-Value Explanations $(M=3.28, SD=1.01)$ and Heatmap Explanations$(M=3.0, SD=1.37)$ in the dimension of understandability ($F(3, 81)= 5.572, p < 0.05$). 
\subsection{Qualitative Analysis and Findings} \label{Exp1_QualAnalysis}
\begin{figure} [H]
    \includegraphics[page=2,width=0.9\columnwidth,trim={0 5cm 0 0},clip]{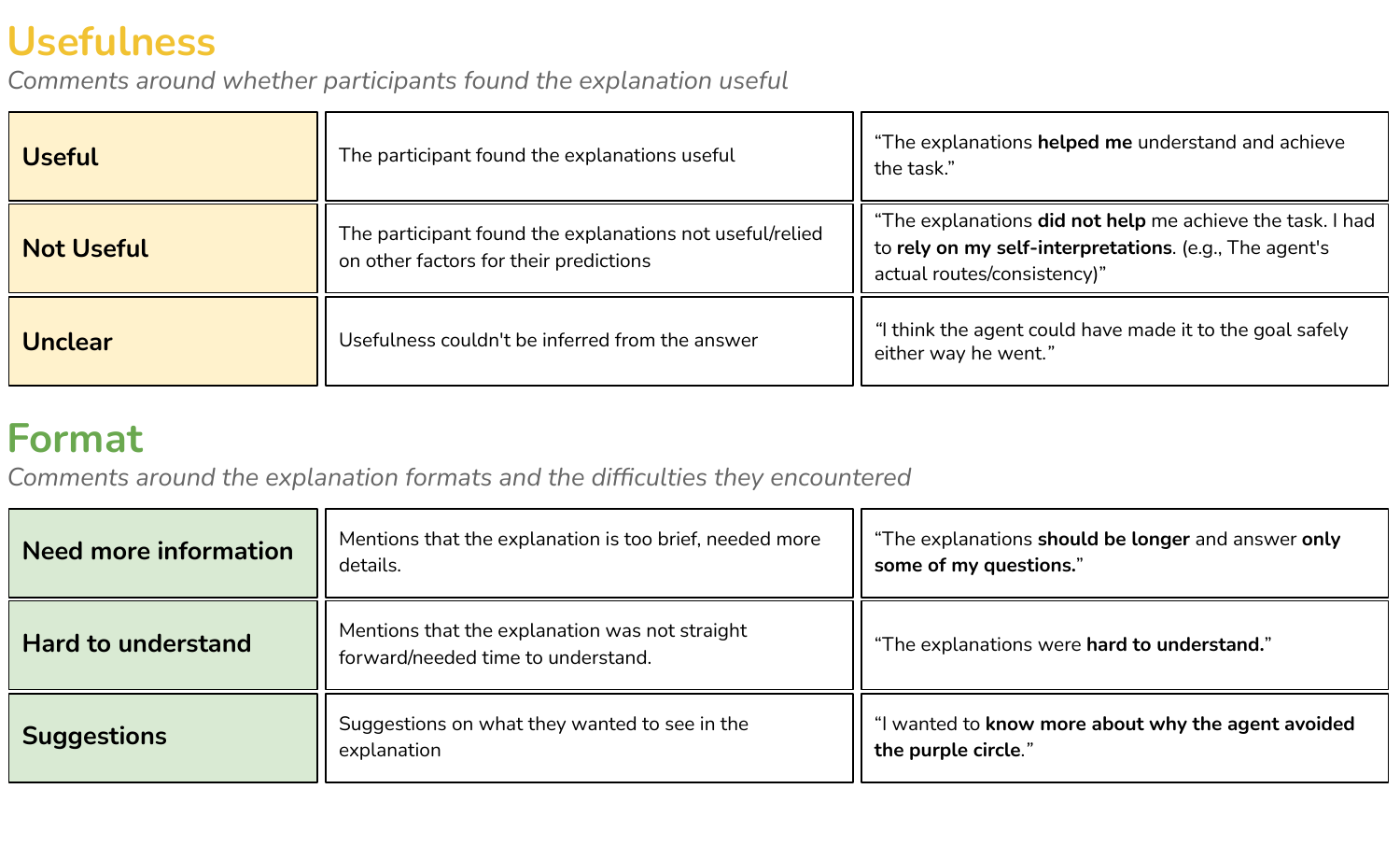}
    \caption{Summary tables of the codes used to understand how users across the explanation groups precieved the explanations in terms of usefulness and format.}
    \label{fig:Exp1_UF_Codes}
\end{figure}

\begin{figure} [H]
    \includegraphics[page=1,width=0.9\columnwidth,trim={0 7cm 0 0},clip]{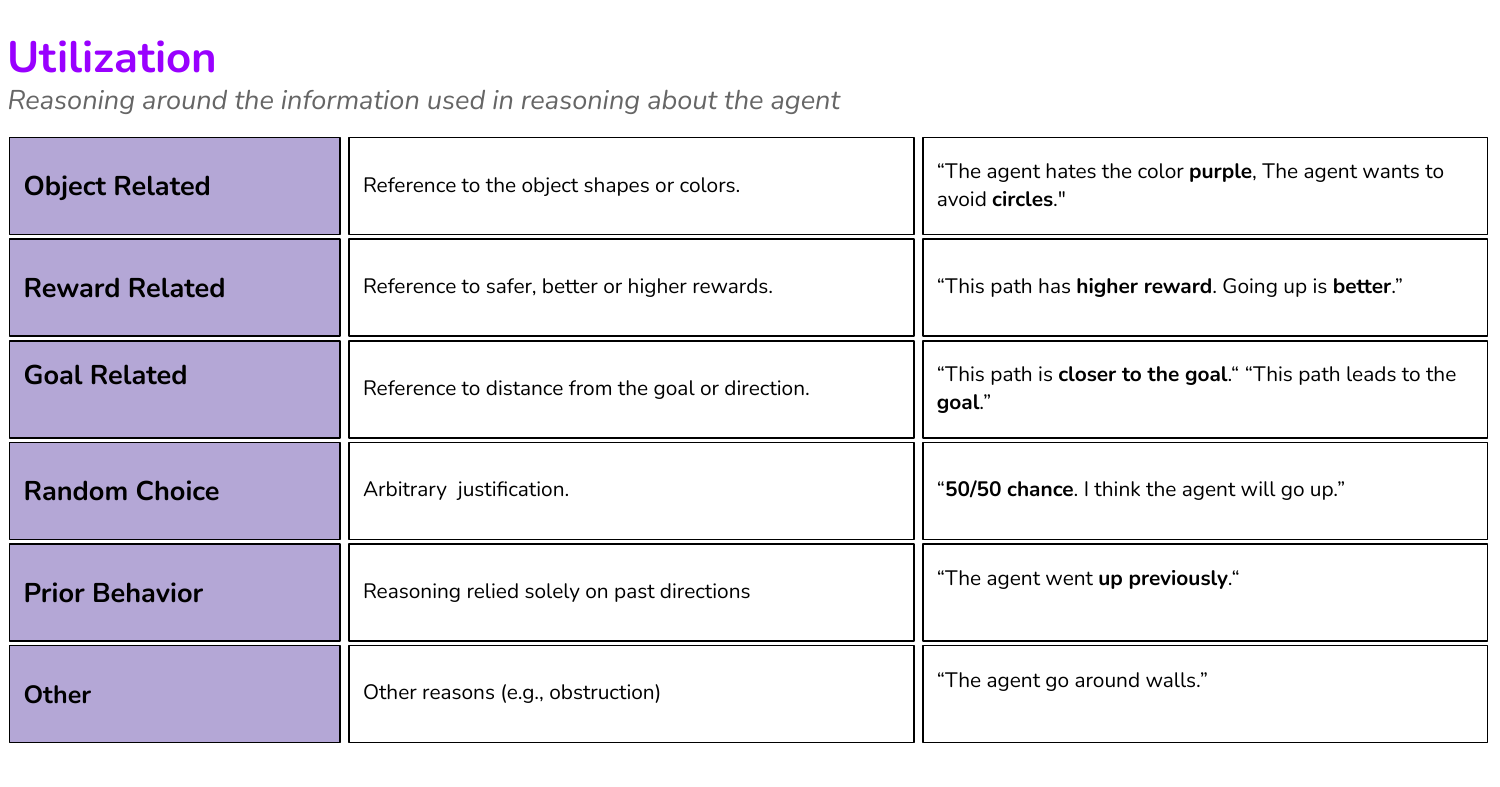}
    \caption{Summary table of the codes used to understand how users across the explanation groups utilized information from the explanations.}
    \label{fig:Exp1_U_Codes}
\end{figure}

To address \textbf{RQ4}, we thematically analyzed open response questions for the tasks and satisfaction survey to better understand how participants used the explanations. Thematic analysis is one of the most common qualitative research methods and involves identifying, analyzing, and reporting patterned responses or meanings (i.e. {\em themes})~\cite{braun2006using}. It is most appropriate for understanding a set of experiences, thoughts, or behaviors across data~\cite{miles2018qualitative}. To conduct the thematic analysis for each of these codes, three authors performed the following established process, as outlined by \cite{braun2006using}: 
(1)~We familiarized ourselves with the data through repeated active readings. 
(2)~We developed a coding framework consisting of a set of {\em {codes}} organizing the data at a granular level and the corresponding well-defined and distinct definitions. 
(3)~Each of the authors independently coded the entirety of the data. If a code was unclear or a new code was needed, the authors consulted with each other to modify the coding framework. If a change was made, each author independently re-analyzed the entirety of the data. 
(4)~We jointly examined the codes to derive themes and findings. 
We compared our application of the codes for each response to ensure consistency. Discrepancies were counted and used to calculate the inter-rater reliability.
The process iterated until all discrepancies were resolved.
We identified three themes: Usefulness, Format, and Utilization. The codes, definitions, and examples are summarized in Figures \ref{fig:Exp1_UF_Codes} and \ref{fig:Exp1_U_Codes}.


\myparagraph{Explanation Usefulness.}
The theme of Usefulness emerged in the open-ended comments about how participants perceived the explanations. Because these comments were made alongside users' explanations of their reasoning, this provides additional information on how useful users found the explanations. We identified
three Usefulness codes: \textcode{olive}{Useful}, \textcode{olive}{Not Useful}, or \textcode{olive}{Unclear} 
We coded a response \textcode{olive}{Useful} if a participant explicitly mentioned that the explanation was useful, for example, ``I understood that purple was dangerous and that the agent should avoid getting near it (EE8).'' 
Responses were coded \textcode{olive}{Not Useful} if a participant explicitly stated that the explanation was not useful, for example,  ``I tried to read the chart but found it confusing and unclear. I didn't understand it very well
(HE10).''. 
Responses were coded \textcode{olive}{Unclear} if the user's perception was ambiguous, for instance, ``I found the explanations a bit too brief. For instance, what makes the purple circles feel safer than the blue squares (EE7)?'' 
The initial inter-rater reliability for Usefulness was 81.5\%.

Table \ref{tab:usefulness-percentage} shows participants' perceived Usefulness across explanation types according to usefulness codes.
Experiential Explanations were considered \textcode{olive}{Useful} most often. Heatmap explanations were considered \textcode{olive}{Not Useful} most often, even more than when No Explanation was provided. A logistic regression comparing the Usefulness of each explanation showed Experiential Explanations were 9 times more likely to be found \textcode{olive}{Useful} than Heatmap explanations and 4 times more likely than Q-value explanations ($p < 0.05$, $df = 3$, $\chi^2 = 14.99$). We also note that 88.89\%  of the participants who thought the explanations were useful predicted correctly in both questions, indicating the correlation between correctness and usefulness ($p < 0.05$, $df = 2$, $\chi^2 = 9.33$).

\myparagraph{Explanation Format.}
Participant responses were also coded for Format. We derived three codes for Format: \textcode{teal}{Need More Information}, \textcode{teal}{Hard to Understand}, and \textcode{teal}{Suggestion}. Responses coded as \textcode{teal}{Need More Information} explicitly stated why the explanation was lacking and what data was desired. If the response said that the explanation was not clear, it was coded as \textcode{teal}{Hard to Understand}, and if the response offered another way to explain, it was coded as \textcode{teal}{Suggestion}. Only responses from individuals who received explanations and referenced the explanation's content could be coded for Format,41.98\% of our dataset; therefore, our reported observations are mainly comparisons of the percentages and could not be statistically tested due to the small data size. Participants often expressed a \textcode{teal}{Hard to Understand} $40\%$ when shown Heatmap explanations, followed by $29.41\%$ who got Q-value explanations while this was not a problem for the Experiential Explanation group. On the other hand, we observed \textcode{teal}{Need More Information} between the three explanation groups with higher concentrations of  $35.29\%$ being from the Q-value explanation group and $26.67\%$ are from the Heatmap explanation groups and  $21.05\%$ are from the Experiential Explanation group. Participants who did not find the explanations useful had more Format comments, especially requesting more \textcode{teal}{Need More Information} $30.77\%$ and \textcode{teal}{Hard to Understand} $69.23\%$. The initial inter-rater reliability was 76.47\%.

\begin{table}[t]
\centering
\footnotesize
\begin{tabular}{lccc}
\toprule
&  Useful  & Unclear & Not Useful \\
\midrule
Experiential Exp.    &        \textbf{80.95\%} &   19.05\% &            0.00\%\\
Heatmap  &       31.58\% &       31.58\% &         36.84\% \\
Q-value Exp. &       47.62\% &       14.29\% &         38.10\% \\
No Explanation      &        N/A & N/A & N/A\\
\bottomrule
\end{tabular}
    \caption{Participant's rates of usefulness for each explanation type.}
    \label{tab:usefulness-percentage}
\end{table}

\begin{figure}[!t]
    \centering
    \includegraphics[width=0.9\columnwidth, height=0.8\columnwidth]{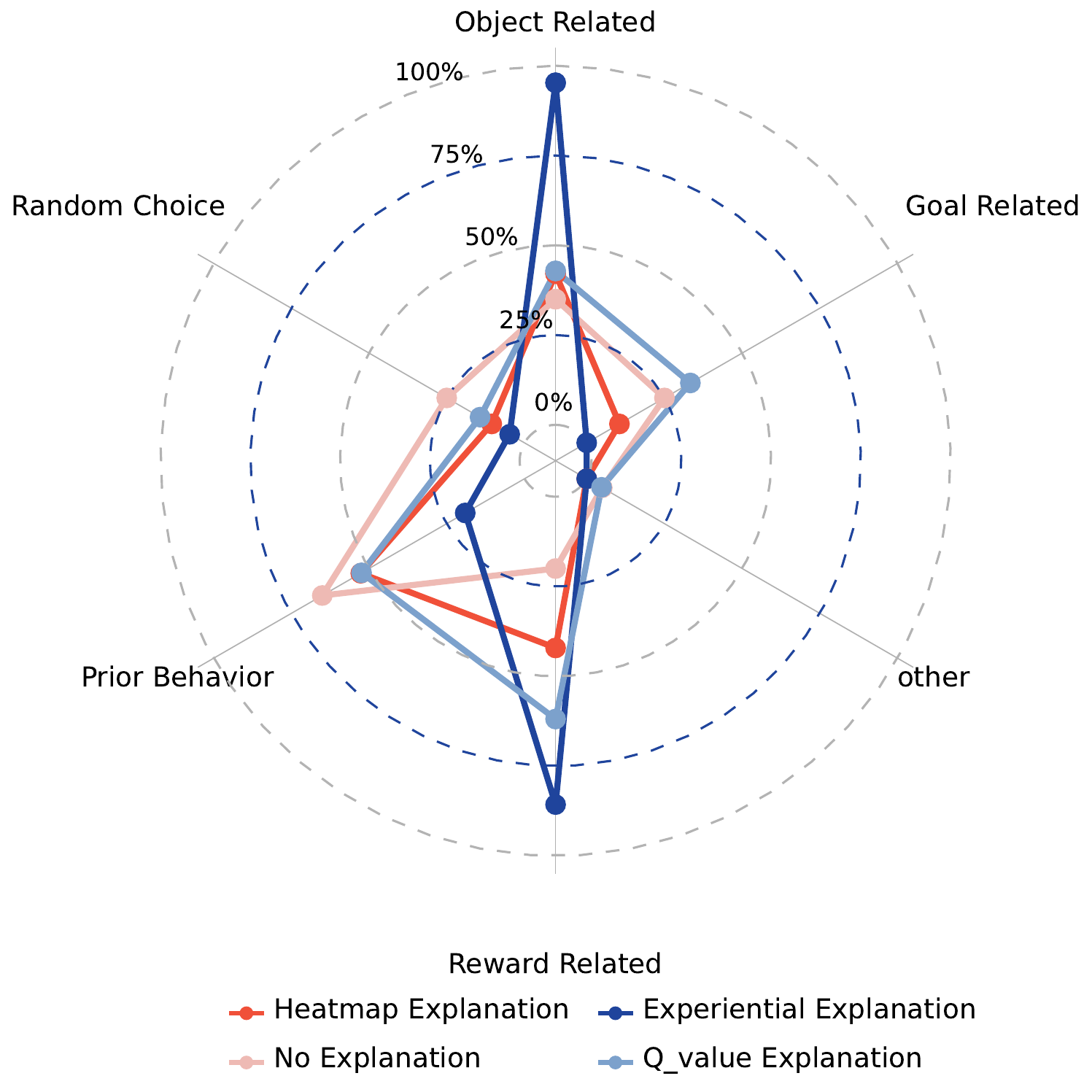}
    \caption{The distribution of themes over the different explanation types. Participants who saw Experiential Explanations, which presented \texttt{Object Related} and \texttt{Reward Related} information, relied more heavily on the kinds of information it provided than other explanation types. Q-Value and Heatmap explanations both presented \texttt{Reward Related} information.
    The breakdown of themes for the heatmap and q-value explanations more closely resemble the no-explanation case.}
    \label{fig:qual-themes}
\end{figure}%

\myparagraph{Explanation Utilization.}
Utilization refers to whether the participant made use of the information available within the explanation to reason about what the agent will do.
We extracted the following codes for the types of information participants referenced in their rationales:
\textcode{violet}{Object Related} (e.g., shapes or colors);
\textcode{violet}{Reward Related} (e.g., ``safer'', ``higher'', or ``lower'' rewards);
\textcode{violet}{Goal Related} (e.g., distance / direction to goal);
\textcode{violet}{Prior Behavior} (e.g., references behavior from prior outcomes);
\textcode{violet}{Random Choice} (e.g., no clear justification);
and
\textcode{violet}{Other} (e.g., references other factors such as obstructions).
The inter-rater reliability of the initial coding was 75.6\%.

We compared the frequency with which utilization codes appeared in open-responses.
If a participant used the same types of information as the explanation provided, we considered the participant as ``Utilizing'' the explanation. 
Experiential Explanations provide \textcode{violet}{Object} and \textcode{violet}{Reward Related} information.
Heatmap and Q-Value Explanations provide \textcode{violet}{Reward Related} Explanations. 
No Explanation only provides for \textcode{violet}{Prior Behavior}.
71\% of the participants who were shown Experiential Explanations utilized the exact kinds of information presented in the explanation, compared to 21\% and 28\% for the Heatmap and Q-Value explanations, respectively.
Although the Q-Value and Heatmap explanations both presented \textcode{violet}{Reward Related} information, participants still relied heavily on \textcode{violet}{Prior Behavior} to understand the agent.

Figure~\ref{fig:qual-themes} shows the distribution of the reasons the participants referenced for each type of explanation. Participants who saw Experiential Explanations most used \textcode{violet}{Object related} and \textcode{violet}{Reward Related} information.  Participants who saw Q-Value explanations referenced \textcode{violet}{Goal Related} reasons the most. Participants who saw no explanation most often utilized \textcode{violet}{Prior Behavior}, followed by \textcode{violet}{Object Related} reasons. 
Notably, participants who saw Experiential Explanations were 4 times {\em less} likely to rely on \textcode{violet}{Prior Behavior}, indicating they did not rely as much on watching the agent to understand it.
Those who saw Heatmap and Q-Value Explanations relied heavily on \textcode{violet}{Prior Behavior} and Utilization patterns mimicking those of the No Explanation condition.

Experiential Explanations were 13 and 15 times more likely to have the information they provided be {\em Utilized} and {\em Useful}, respectively.
{\em Utilized} refers to what they used in their reasoning, while {\em Useful} focuses on whether they thought the explanations were useful or not in the post-task satisfaction survey. Yes, it is the code from the previous section. with respect to Heatmap and Q-Value explanations, respectively.  We also found that among those who used components of explanations in their reasoning, 55.88\% answered correctly on both questions ($p < 0.05, df=1, \chi^2 = 3.849$). But within the Experiential Explanations group, the percentage was 84.62\% and that group was more statistically significant to get both questions correctly than other baseline explanation groups ($p < 0.05, df=3, \chi^2 = 12.22$).

\section{Experiment 2: Crafter Environment}

In this experiment, we address our research questions (RQ2, RQ3, and RQ4) with a focus on evaluating the actionability of the explanations in a more complex environment, as this experiment is designed to evaluate the Influence Predictors and Experiential Explanations in a visually richer environment, with more decision points and options, along with multiple reward classes. And, as before, we will validate the faithfulness through automated testing designed to address (RQ1). 

To create such a setting, we used the Crafter environment developed by \citeauthor{hafner2021crafter} \cite{hafner2021crafter}, a 2D version of the game Minecraft with the same rules for interacting with the environment and crafting new tools.
Crafter can generate 2D open-world survival environments with randomized maps. These environments consist of varying terrains, a day-night cycle, as well as interactable materials, objects, and living entities. Crafter also supports an on-screen inventory, allowing an agent to collect items in order to construct a variety of items later. An agent playing Crafter needs to complete various tasks in sequential order while maintaining its health by obtaining resources and avoiding threats. 

\subsection{Model Setup and Environment Details}
\label{sec:crafter_method}

\begin{figure}
\centering
    \includegraphics[width=\columnwidth]{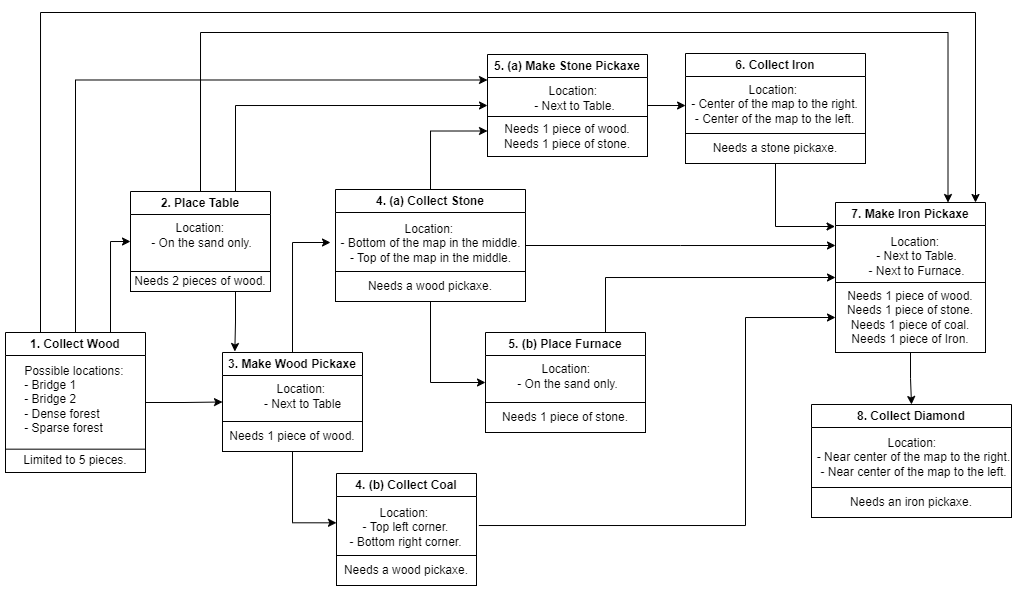}
    \caption{Depiction of the task dependencies of the Digger's quest, listing the achievements that the Digger needs to attain and the order in which that needs to be done}
    \label{fig:digger_direct_sequence}
\end{figure}

\myparagraph{Environment Customization.}
We needed a good distribution of positive and negative reward sources. Crafter provides multiple sources of positive rewards, but the only negative rewards come from health depletions. 
To counter this and to add a conceptual narrative, we created an agent persona, called the Digger, whose aim is to obtain a diamond. 
The Digger's reward system provides positive rewards for achievements related to digging, such as acquiring resources to construct digging tools, creating these tools, and digging for minerals. We provide negative rewards if the agent strays from the Digger persona, for example, by constructing weapons, or killing an entity. Figure \ref{fig:digger_direct_sequence} depicts the sequence of achievements required to reach a diamond.

Additional customizations were added to optimize the training process. The world map was made static, so that the agent could find all required resources in a limited number of steps while making it easier for a participant to understand the world map. No rewards were given for performing actions to maintain health, such as drinking water, but improvement or depletion of health had correponding positive and negative rewards. While all other achievement rewards are in the range of +1 to -1, diamond collection has a +50 reward to focus the agent's training exploration towards getting a diamond. Finally, a small bonus is given for reaching the diamond in fewer timesteps, to improve agent performance. 

\myparagraph{Agent training.}
The agent needs to learn to traverse the map, completing the achievements required to reach the diamond in a small number of timesteps. At any given point, its observation consists of an image of its surroundings inside a limited field of view along with its current inventory. Given a static map, the agent is trained from various starting points to aid in the training exploration. We use on-policy learning algorithms for the agent, advantage actor-critic (A2C)~\cite{willems-2018} 
optimized using the exponential policy optimization algorithm (PPO) \cite{schulman2017proximal}, implementation details for which are available in Appendix \ref{appendix:crafteragent}.

\myparagraph{Influence Predictor design and training.}
For the Crafter experiment, we attached Influence Predictors (IP) to achievement categories. Each predictor was responsible for tracking either favorable achievements leading to positive rewards or unfavorable achievements leading to negative rewards. The Health-tracking IP was the only exception here, as it tracked both improvement and depletion of health, resulting in rewards that are sometimes positive and sometimes negative. 
Table \ref{tab:influence-predictors} lists all Influence Predictors, along with the achievements they track. The Health tracking influence predictor is a special case of a reward signal that is both negative and positive. We opted not to divide the health rewards coming from the environment, because we didn't want to change the health signal artificially; we would need to significantly intervene to seperate the signal into its positive and negative components. To handle this special case we don't use the absolute function for the health tracking influence predictor; we took both the sign and the magnitude of the values when it came to the health IP. This still works because we still standardized all the values before the explanation generation which corrects any inconsistency this inclusion of the sign might have caused. 
\begin{table}[t]
    \centering
    \footnotesize
    \begin{tabular}{lllc c c}
        \toprule
        Influence Predictor & Reward type & Achievements \\
        \midrule
            Coal collection & Positive & collect-coal\\
            Diamond collection & Positive & collect-diamond\\
            Iron collection & Positive & collect-iron\\
            Stone collection & Positive & collect-stone\\
            Wood collection & Positive & collect-wood\\
            Iron pickaxe construction & Positive & make-iron-pickaxe\\
            Stone pickaxe construction & Positive & make-stone-pickaxe\\
            Wood pickaxe construction & Positive & make-wood-pickaxe\\
            Furnace placement & Positive & place-furnace\\
            Table placement & Positive & place-table\\
            Murder tracking & Negative & defeat-zombie\\
            Iron sword construction & Negative & make-iron-sword\\
            Stone sword construction & Negative & make-stone-sword\\
            Wood sword construction & Negative & make-wood-sword\\
            Health tracking & Positive and Negative & life-maintenance, collect-drink, eat-cow\\
        \bottomrule
    \end{tabular}
    \caption{Complete list of Influence Predictors with the achievements that they track and their corresponding reward types}
    \label{tab:influence-predictors}
\end{table}

The Influence Predictors were also trained using on-policy learning, the technical details can be found in Appendix \ref{appendix:crafteragent}. The predictors were given full access to the agent's rollout buffer allowing them to see the agent's observation space, action taken, and reward obtained, along with whether the episode just started and the probability distribution of agent's actions. IPs would only read the rewards pertaining to their own achievements and consider rewards from other sources as 0. Also, instead of taking the actual reward, IPs read the magnitude of the reward to track the intensity of influence of their achievements, irrespective of whether they are positive or negative. The Health-tracking IP would also consider the sign of its reward, given that health can be both positive and negative. The predictor models calculated the value and subsequently the advantage and results using their own respective critics. They would then populate their own rollout buffers with these details and train on the same. Each IP hence would learn its individual policy based on the agent's explorations.

\begin{figure}
    \includegraphics[width=\columnwidth]{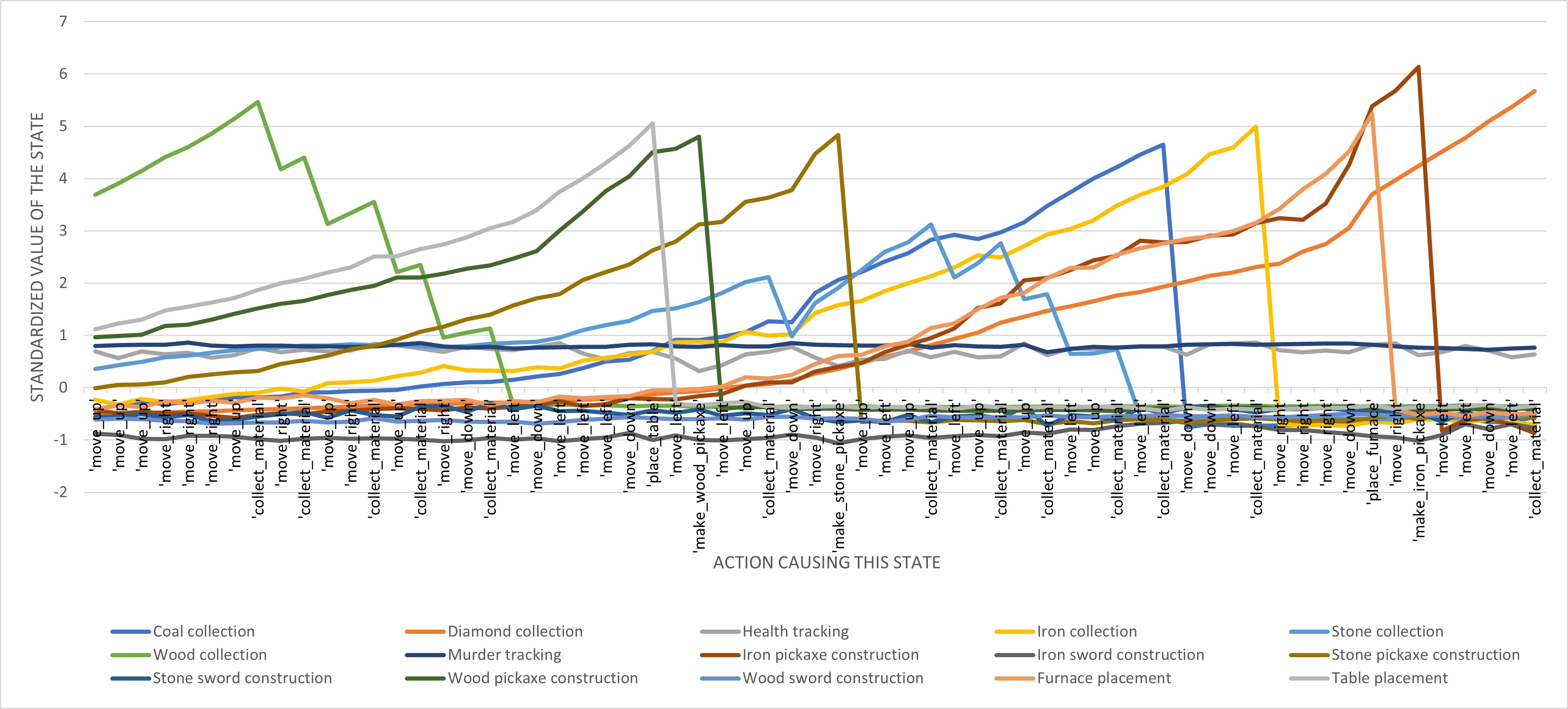}
    \caption{A plot of an agent's full trajectory over an episode and the standardized values of every state along the trajectory according to individual Influence Predictors. The $x$ axis shows the temporally ordered actions preceding each state whose influence predictor values are represented on the $y$ axis.}
    \label{fig:stan-values-example}
\end{figure}

We explored the IP values by standardizing them to compensate for reward variations and obtained graphs such as Figure \ref{fig:stan-values-example}. We see that the influence of rewarded states associated with achievements keep rising until the achievement was obtained and drop immediately after, leading to creation of spikes as visual indicators of when an achievement was made. The IPs hence expose the user to knowledge about the agent's goal hierarchy because the positive rewards are attached to achievements. This is something that we don't see in the simpler minigrid and a useful emergent property of the technique.

\myparagraph{Explanation generation.} To explain the actions of the agent, we provide local explanations that narrate what the agent plans to do next after the actual or counterfactual action, as described in Section \ref{sec:explanation_generation} and using the following template. \\
\textbf{Template:} \textit{``[The agent] needed [highest influence predictor at decision point] and chose to [$a_\mathrm{agent}$]. This led to the prioritization of [list of top influences for $t_\mathrm{agent}$] next. However, if it had [$a_\mathrm{user}$], it would have to think about [list of top influences for 
 $t_\mathrm{user}$] instead.''} \\
\textbf{Example:} \textit{``Jade needed wood and chose to get it from the sparse forest. This led to the prioritization of building a table, building a stone pickaxe, and getting stones next. However, if it had gotten to the dense forest, it would have to think about owning a wooden pickaxe, fearing for its health and avoiding dangerous zombies instead.''} \\
This explanation conveys that the dense forest is a more dangerous option for the agent, because the agent knows, while the user might not, that there are zombies roaming around that forest and it would be wiser to get wood from the safer sparse forest. We anthorpomorphize our agent; calling it ``Jade'' in all explanation groups and make minor edits for readability as needed (e.g., change the tense or add articles) to improve the survey experience of the participants.

\subsection{Faithfulness Evaluation Methodology}

\begin{figure}
    \includegraphics[width=\columnwidth]{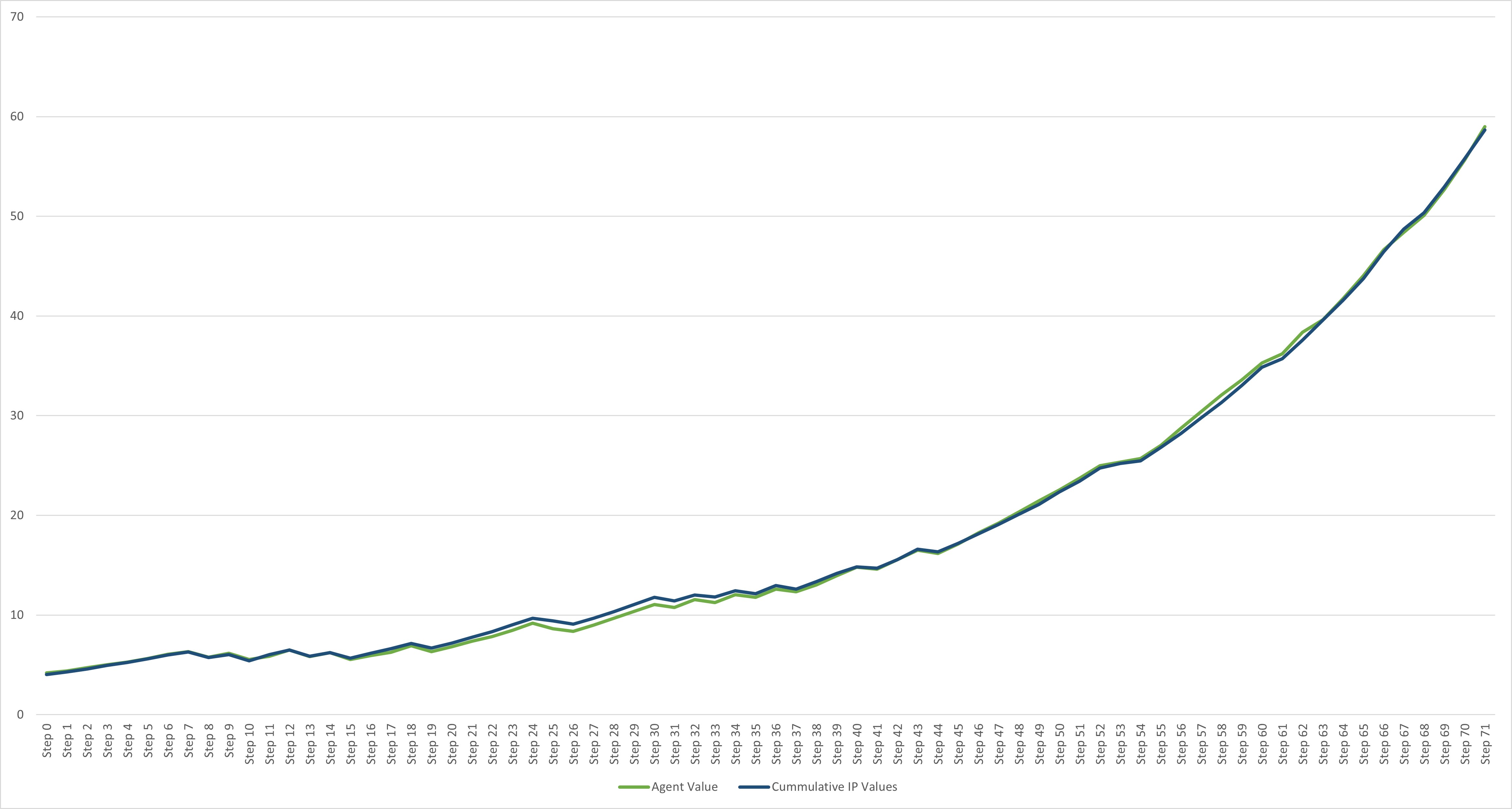}
    \caption{This plot depicts the values of different states as traversed by the agent during a successful test run. The green line represents the values according to the agent's learned policy. The deep blue line depicts the cumulative values according to the policies learned by the Influence Predictors. Both plots match almost exactly, meaning that the policies learned by the Influence Predictors align closely with the policy learned by the agent. }
    \label{fig:faithfulness-example}
\end{figure}
We look at faithfulness to evaluate how good our explanation method is at approximating the agent. We calculate faithfulness (RQ1), which looks at how the influence predictors $(IP)$ models can align with the agent's model. We used the two faithfunlness evaluation methods from the first experiment as described in Section \ref{faith_method} to estimate an agent's actions based on influence predictors. The first method combines the values of all the positive and negative influence predictors to calculate a cumulative value, allowing us to determine the agent's action as follows: $[ \text{action} = \text{argmax}_a \left( \sum_i P_i(s, a) - \sum_j N_j(s, a) \right) ]$
where $( P_i )$ represents a positive influence predictor and $( N_j )$ represents a negative influence predictor.

The second method involves training a classifier using the standardized positive and negative influence values as features, with the agent's actions as the ground truth. We used the same setup we had in the first experiment, an SVM classifier with an RBF kernel.

To learn more about how our influences perform in different parts of the trajectory, we take a slightly different approach than the one we did in the first study. In the first study we did so by focusing on how the influences values impact the agent at a close proximity. In this experiment, because our setting here is more complex and combining the Influence Predictors indvidiual action predictions can be noisy, and it is not feasible to easily identify the strengths of their signals at different locations, as not all the 15 IPs are equally influential at all decision points we resort to looking at the sum of their values at each state as a way to show their alignment with the agent's. Then we compare this sum with the agent's values for each state. The goal is to see how closely the cumulative IP values can match those of the agent. Our process begins by obtaining the cumulative IP values of each state using the formula:
\begin{equation}
\label{eq:sum_IP}
\text{Cumulative IP value for state s} = \sum_{IP_c \in C^{+}}{IP_c(s)} - \sum_{IP_c \in C^{-}} {IP_c(s)}
\end{equation}
Where $C^{+}$ is the set of positive IPs including Health tracking IP and $C^{-}$ is the set of negative IPs.

We then calculate the Root Mean Squared Percentage Error (RMSPE) \cite{CHAN1995135} between the agent values and the cumulative IP values for a set of 9 test runs. These runs used the same trained agent on the same map, but varied the starting location. 
RMSPE is calculated as: 
\begin{equation}
RMSPE = \sqrt{ \frac{1}{N} \sum_{n=1}^N{ \left( \frac{\hat{y}_n - y_n}{y_n} \right)} ^2
}
\end{equation}

As before, the faithfulness evaluation does not use the baselines as the purpose is to evaluate the influence predictor models' ability to approximate the policy model.

\subsection{Baselines}
\label{sec:exp2_baselines}
The setup of this experiment allows the generation of localized explanations, so we used the following baselines to compare them with:
\begin{enumerate}
\item \textbf{Local Agent-Value explanation:} For both options provided to the user, this explanation presents the agent value of the state reached right as the action sequence ends. This action sequence will end in some reward-producing state, but not necessarily in the goal state. Similarly to the previous experiment, these values are generated using the simulator. We similarly used templates to generate the explanations encapsulating the values.
\\\textbf{Template:}\textit{Jade's value for [$a_\mathrm{agent}$] is [agent-value after $a_\mathrm{agent}$]. While the Jade's value for [$a_\mathrm{user}$] is [agent-value after $a_\mathrm{user}$]}
\\\textbf{Example:}\textit{Jade's value for getting wood from the sparse forest is 6.99. While Jade's value for getting wood from dense forest is 5.59}
\item \textbf{No explanation:} As in the first experiment, users only saw the correct answer, without further explanation.
\end{enumerate}
The agent-value explanation is similar to the Q-value explanation we had in the previous study, except now it is calculated on a local point rather than an aggregate mean over the full trajectory. 
We did not include the heatmap baseline because the agent may need to traverse the same positions on the map multiple times to complete its task, which made heatmaps inapplicable.

\subsection{User Study Methodology}
We follow a study design similar to our first experiment; We have similar main study components of prediction tasks (RQ1, RQ3) and a satisfaction survey (RQ2, RQ3) as in our previous study. We adjusted the prediction task to be more suitable for the new environment and local explanations. We also introduce a new ordering task to further explore the comprehension of the explanations that address RQ3 and RQ4. 

Our study is an online evaluation between subjects using Prolific with 96 participants aged 18-82 years old (M=38.90, SD=12.91); 32.65\% of the participants identified as women. The median study completion time is approximately 12 minutes, and each participant was compensated with \$2.50 with a bonus for thoughtful responses.

Initially, to introduce the participants to the map and receive the initial explanation, we show them a scenario and then ask them what the agent would choose between two options. After making a selection, they will see an explanation generated by our technique or one of the baselines. The explanation group selection were fully randomized and balanced using Qualtrics.

Then they will see another scenario, {\bf a prediction task}, where they are asked to predict the agent's follow-up action; they are presented with one of two prediction tasks. The first examines the case of multiple positive influences, where they are asked which action will the agent take next. The other examines the negative influences, where they are asked which option the agent will avoid. In both versions, the correct options relate to the influence with the highest value. After making their predictions, we ask them to explain their reasoning. 

The next task is {\bf an ordering task}, where we show them a list of the factors affecting the agent for each of the action choices of the initial scenario. And ask them to order them according to their relative importance to the agent. We also ask them what factors they considered with the ordering. Then we ask if the explanations were helpful in this task and how.

Finally, they conduct {\bf a satisfaction survey}, identical to experiment 1, where they rate the explanation on understandability, satisfaction, amount of details, completeness, usefulness, precision, and trust. In addition, a text entry question about the usefulness of the explanations to their decision making. We provide a detailed description of the survey and a screenshot in the Appendix \ref{appendix:eval-survey2}.

\begin{figure}
\centering
\includegraphics[width=0.49\columnwidth]{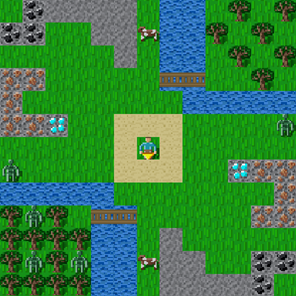}
\includegraphics[width=0.49\columnwidth]{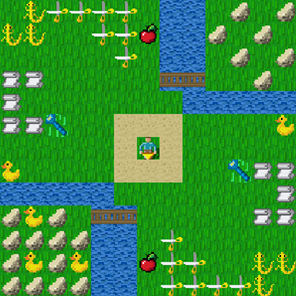}
\caption{Comparison of a 13x13 tile Crafter map with original art assets on the left and re-skinned art assets on the right}
\label{fig:crafter-reskin-comparison}
\end{figure}

\myparagraph{Disguising of assets}
The functionality of Crafter's materials, objects, and living entities is quite intuitive, and many people are familiar to different degrees with Minecraft. 
We needed to ensure that the explanations were the only source of a participant's understanding of concepts of this world, as opposed to prior knowledge about Minecraft or commonsense. 
We re-skinned the art assets to random alternatives which carry no relation to the thing they represented. Figure \ref{fig:crafter-reskin-comparison} displays the main map with a subset of original and re-skinned assets. The full list of re-skinned assets can be found in the technical appendix.

\subsection{Quantitative Results}

\myparagraph{RQ1: Faithfulness}
For the first test, the accuracy of the influence predictors in predicing the agent's actions along the trajectory, we found that the accuracy of the prediction of the direct aggregation to be $73.35\%$.

Using the classifier from experiment 1 with only movement actions (4 classes) resulted in $69.61\%$ and with all possible actions (11 classes) is $60.71\%$.
The classifer from experiment 1 only uses influence predictor values as features, not the entire observation; this was sufficient for experiment 1.
However, the Crafter environment is much more complex in terms of action space, state space, and temporality---states are seen repeatedly but under different conditions such as different tasks.
Furthermore, the size and complexity of the state-action space is such that certain trajectories through state space are vastly over-sampled by the policy learner while others are under-sampled.
This makes it hard to learn a classifier without artificially forcing a more balanced sampling of states and actions.
Other classifier architectures may perform better. 
Regardless, it shows that the influence predictors provide information that can be used to reconstruct the policy many times above random chance ($9\%$) even using naive reconstruction means such as this classifier.

The cumulative values of the influence predictors, calculated with equation \ref{eq:sum_IP} were almost identical to those of the agent.  We plot the results of an example test run of average length that shows how the values are aligned almost perfectly, as demonstrated in Figure \ref{fig:faithfulness-example}. This is verified by the RMSPE values that averaged 3.13\% $(SD=0.52\%, min=2.45\%, max=3.82\%)$, indicating a high similarity between the agent policy and the cumulative IP values.

\myparagraph{RQ2: Prediction Correctness and Order Correctness} We measure whether participants can correctly predict the agent's action at the prediction task decision point after interacting with the explanation of the initial decision point. The correctness rose up for the local Experiential Explanation group from $0.4 \rightarrow 0.71$, while there was no difference in the other groups as shown in table \ref{tab:prediction_task_correctness}
The group that received local experiental explanations demonstrated a significant improvement in the accuracy of the prediction in all groups after exposure to an explanation. The results of the logistic regression model were statistically significant $(X^2(2, N = 96) = 13.327, p < 0.05)$. There was a positive correlation between the local Experiential Explanation group and the accurate prediction of influences $(OR: 2.55; 95\% CI [1.785, 3.326])$.

\begin{table}[]
\begin{tabular}{@{}lll@{}}
\toprule & \textbf{Initial Predictions} & \textbf{Prediction Task}               \\ \midrule
\multicolumn{1}{l}{\textbf{Experiential Explanation}} & \multicolumn{1}{l}{40.63\%} & \multicolumn{1}{l}{\textbf{71.88\%*}} \\ \midrule
\multicolumn{1}{l}{\textbf{Agent-value explanation}}  & \multicolumn{1}{l}{28.13\%} & \multicolumn{1}{l}{28.13\%}           \\ \midrule
\textbf{No explanation}                                 & 37.50\%                      & 34.38\%                                \\ \bottomrule
\end{tabular}
\caption{A comparison of the explanation groups prediction correctness for their initial predictions; before seeing any explanations and after seeing the relevant explanation.}
\label{tab:prediction_task_correctness}
\end{table}

\begin{figure}
    \centering
    \includegraphics[width=\textwidth]{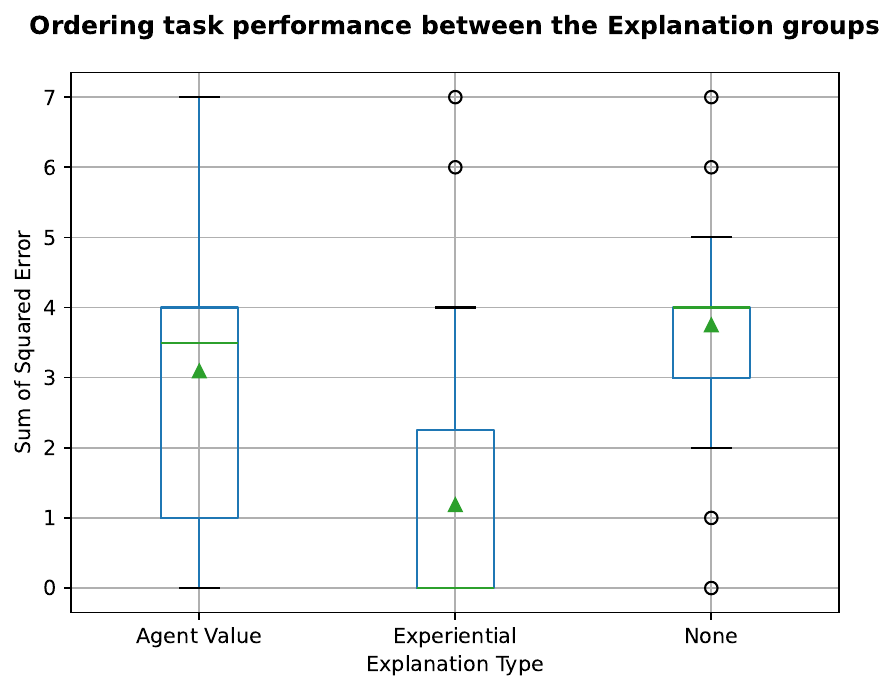}
    \caption{A comparison showing the distribution of the sum of squared errors for the ordering task between the different explanation groups.}
    \label{fig:boxplot-ordering}
\end{figure}

For the ordering task, we measure the ability of the participants to order the factors that impact the agent the most for each action proposed at the initial decision point. We used Sum of Squared Error to compare the correctness of the sequences as shown in Figure \ref{fig:boxplot-ordering}. The group provided with local Experiential Explanations showed significantly reduced squared errors (M=1.19, SD=2.27) compared to other baseline groups which is found by conducting a one-way analysis of variance (ANOVA) followed by Tukey’s honestly significant difference test
 $(F (2, 96) = 14.1315, p < 0.05)$. This indicates that the local Experiential Explanation group was more effective in achieving accurate ordering compared to the agent value explanation group $(M=3.09, SD=2.89)$ and the group without explanations $(M=3.63, SD=2.91)$. 

\myparagraph{RQ3: Satisfaction Survey} As with the first experiment, we analyzed the explanation goodness ratings the participants provided for each explanation group for the Likert scale questions.  We include a visualization of the results in Figure \ref{fig:ex2-survey} in the Appendix. The local Experiential Explanation group exhibits the most favorable mean values in all measured dimensions, with the exception of Trust. A one-way analysis of variance (ANOVA) followed by a Tukey's honestly significant difference test revealed that these differences are not statistically significant, with the sole exception of the dimension of sufficient detail $(F(2,96) = 4.716, p < 0.05)$, where the local Experiential Explanation $(M=3.33, SD=1.45)$ was perceived as significantly more detailed compared to the agent value explanation $(M=4.06, SD=1.16)$.

\subsection{Qualitative Analysis and Findings}
To understand how users in different explanation groups varied in the way they used the information provided in explanations, we thematically analyzed the open response questions asking participants why they selected their particular answers for the reasoning tasks. We followed the same procedure we describe in Section \ref{Exp1_QualAnalysis}, familiarizing ourselves with the data, developing a coding framework, independently coding the data, and jointly resolving differences \cite{braun2006using}. Discrepancies were counted and used to calculate the inter-rater reliability. The process was repeated until all discrepancies were resolved. The inter-rater reliability for all three questions we analyzed are 71\%, 75.1\%, and 80.1\%. 

\begin{figure} [H]
    \includegraphics[page=1,width=0.9\columnwidth]{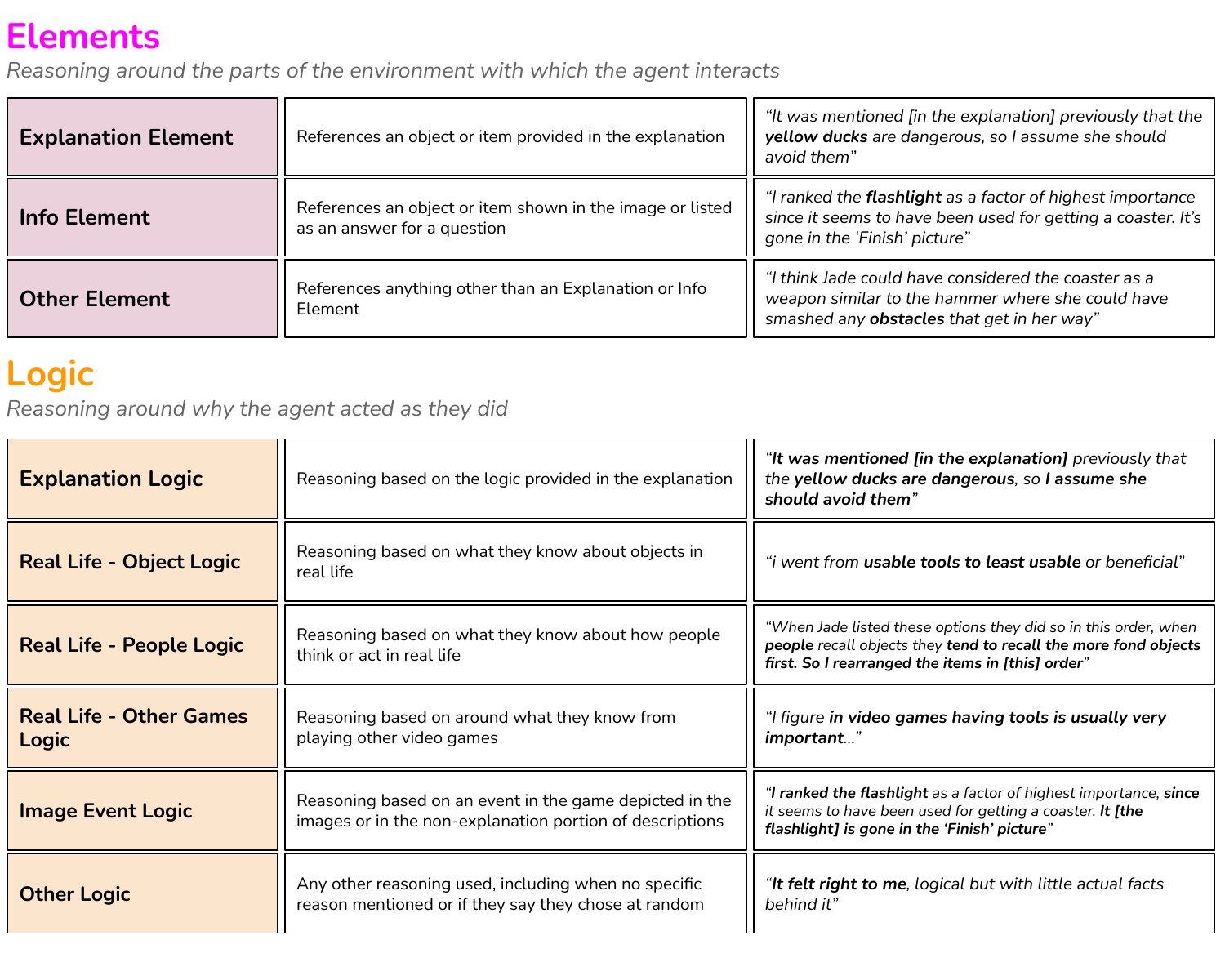}
    \caption{Summary table of the codes used to understand how users across the explanation groups utilized information from the explanations. Element codes refer to parts of the environment that participants thought the agent would interact with. Logic codes refer to participants' beliefs around the circumstances leading to the agent's actions.}
    \label{fig:Exp2_Element-LogicCodes}
\end{figure}

Using this procedure, we identified the following codes that are broadly separated into: (1) \textit{elements}, participants understanding the parts of the environment the agent would interact with; and (2) \textit{logic}, participants reasoning around the circumstances leading to the agent's actions. The codes, definitions, and examples are summarized in Figure \ref{fig:Exp2_Element-LogicCodes}. Significantly, the codes for Experiment 2 reflect and extend the codes identified for Utilization in Experiment 1, but they were developed independently, grounded in each set of data, demonstrating an additional layer of validity and consistency between the two studies' findings.

\begin{figure} 
    \includegraphics[page=2,width=\columnwidth, clip, trim=0cm 8cm 0cm 0cm]{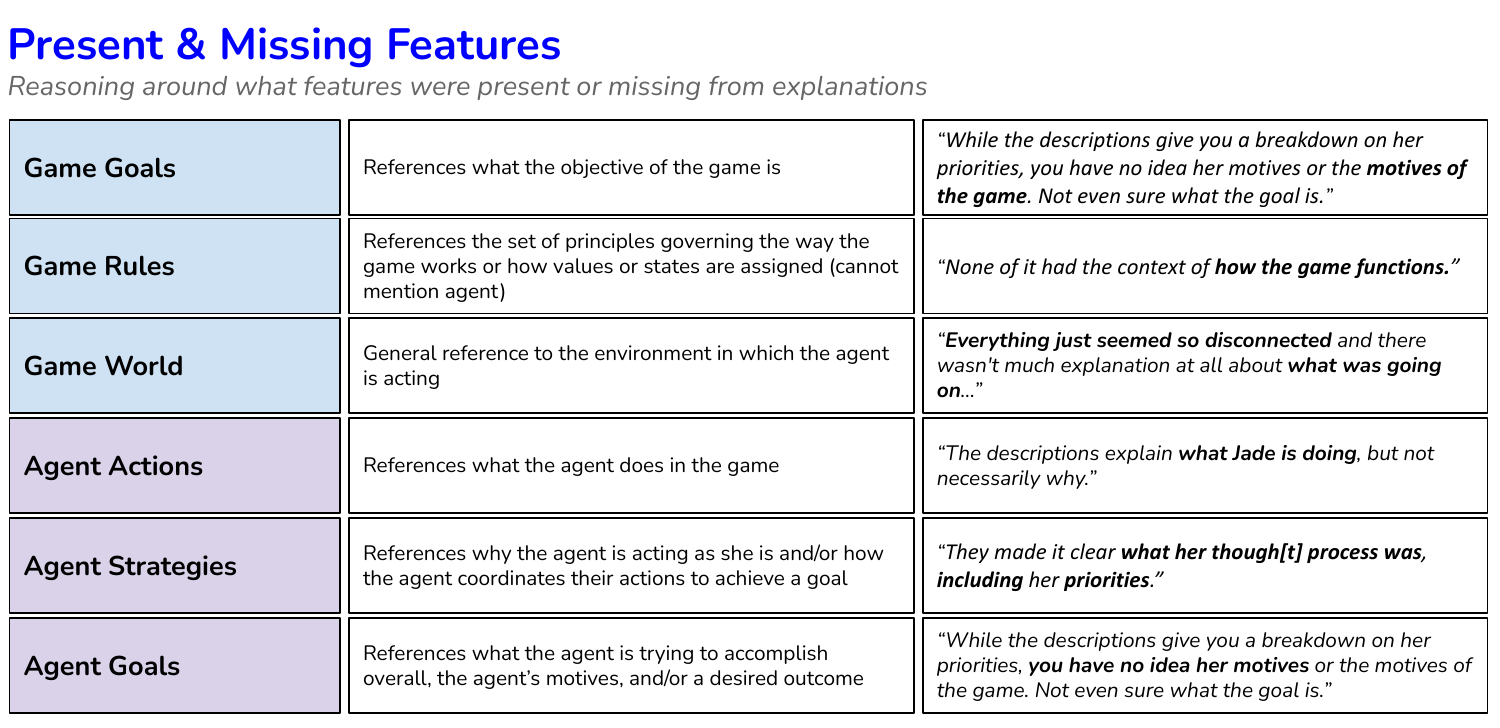}
    \caption{Summary table of the codes used to understand what features users perceived as present and missing from the explanations. Features are separated by if they tie to the game or the agent.}
    \label{fig:Exp2_ExplanationGoalCodes}
\end{figure}

We also used the same procedure to conduct a thematic analysis of the open response question from the satisfaction survey asking participants about the usefulness of the explanations. We qualitatively coded their responses to identify the features that users perceived as present and missing from the explanations. We broadly separated our codes into features related to the broader \textit{Game} (e.g., game rules) and features related to the  \textit{Agent} (e.g., agent strategies). The codes, definitions, and examples are summarized in Figure \ref{fig:Exp2_ExplanationGoalCodes}. The inter-rater reliability for this question is 83.5\%.

We focus our discussion on the most salient themes we identified from our analysis. (1) Participants in the different explanation groups vary in how they utilize the information provided in the explanation. (2) Participants identified the features they perceived as missing from the explanations. (3) In the absence of a complete picture, how do participants reconcile their mental models of the agent? 

\myparagraph{Comparing Elements \& Logic Utilization Across Explanation Groups.}
The group provided with local Experiential Explanations was the only group to demonstrate \texttt{direct reliance}, relying only on the Explanation Logic and Explanation Element with nothing else added. Of the 96 questions answered by the local explanation group, 41 (42.7\%) directly relied on the explanation provided, compared to none in either of the other two baselines. Those who did not directly rely on the explanation most often relied on both the \textcode{orange}{Explanation Logic} and  \textcode{magenta}{Explanation Element}, but added other explanations or elements. The group provided with agent value explanations most heavily relied on  \textcode{orange}{Real Life - Object Logic},  \textcode{orange}{Other Logic}, and  \textcode{magenta}{Info Elements}. The group provided with no explanations most heavily relied on  \textcode{orange}{Real Life - Object Logic},  \textcode{orange}{Image Event Logic}, and  \textcode{magenta}{Info Elements}. This indicates that participants in the local Experiential Explanation group more heavily relied on the information provided in the explanation to predict the agent's actions, rather than drawing on other knowledge. It also confirms the experimental validity, as not all participants in the local Experiential Explanations solely depended on the explanations, indicating answers were not directly given to them in the explanation.

Overall, using influence predictors provided a significant improvement over other explanation methods in helping users predict the agent's downstream actions. Participants who received the local Experiential Explanations were significantly more likely $(p < 0.05)$ to corretly choose the influences on the model compared to the other explanation groups. Further, those who received local Experiential Explanations more frequently relied on the information provided in the explanation, rather than what was provided in the images or on connections to experiences outside of the experiment. 42.7\% of the participants in the local Experiential Explanations group directly relied on only the explanation elements and explanation logic provided to make their decision.

The improvement in performance aligns with the provision of additional information through our method. Influence predictors are models that learn how sources of reward affect the agent in different stages, restoring information about how the policy reflects the environment. This information allows users to predict the downstream actions of the agent as opposed to the other experimental groups that could not provide this information. Hence, those presented with local Experiential Explanations had significantly higher correct predictions and reliance on the elements and logic of the explanation given, compared to baselines. The Agent-value explanations baseline did not have information that could be translated to make downstream predictions, and so those participants relied more heavily on \textcode{orange}{Real Life Object} and \textcode{orange}{Other Logic} to fill in the gaps around what the agent might do. Similarly, participants who were not provided an explanation at all were left in the dark entirely, relying mostly on \textcode{orange}{Image Event Logic}, \textcode{orange}{Real Life Object}, \textcode{orange}{People}, and \textcode{orange}{Other Games Logic}. Further, no participants in either the Agent-value explanations or the no-explantion groups were able to correctly predict the agent's actions and order the agent's priorities as compared to 13 of the 26 local Experiential Explanation participants who were able to correctly do both.

The qualitative analysis of participants' responses to how well they felt they understood the agent revealed implications for the design of explanations in complex environments. Many people in the local Experiential Explanation group directly relied on the explanation and felt they understood the agent. For example, one participant clearly stated what several others echoed, ``They made it clear what her though[t] process was, including her priorities'' (L31)\footnote{The first letter in the participant code indicate their explanation group. (L) Local Experiental Explanations, (A) Agent-Value Explanation and (N) No explanation.}. However, others expressed that they wanted more from the explanations. As one participant said, ``The descriptions explain \textit{what} Jade [the agent] is doing, but \textit{not necessarily why}'' (L28, emphasis added).

\myparagraph{The Multi-Dimensional Nature of Users' Needs in Complex Environments}
The nature of the task is extremely disguised to ensure that participants would not be able to rely on prior knowledge of the Crafter game or commonsense. 
This is important for understanding how much information can be derived from the explanations and put to use in making predictions about the agent and the environment.
The extreme disguise of the task was also useful in that it elicited feedback about what users {\em wanted} regardless of whether they were successful or unsuccesful at the task.
These are mainly reflected in the \textit{Game} (\textcode{blue}{Goals, Rules, World}) and the \textit{Agent} (\textcode{blue}{Actions, Strategies, and Goals}) codes.
These desires were not only expressed in the local Experiential Explanation group, but were also expressed across the Agent and No explanation groups. For example, a participant in the no explanation group who did not get any correct answers said ``There is no context or logic to the decision-making process from the information given'' (N16), which almost exactly mirrored the statement of a local Experiential Explanation participant who correctly responded to all the questions, ``I don't know what she [the agent] chose or what her goal is, so her motivation doesn't make sense to me'' (L19). Despite the ability to perform well on the task, the participant still desired additional information beyond what influence predictors could provide, revealing the multi-dimensional nature of the needed explanations. An explanation of the rewards influences on the agent through influence predictors helps clarify some of the causal dimensions, which were enough to predict the agent's  functions, yet they fall short of providing a comprehensive understanding of the agent's underlying motivations or broader environmental context that remains outside the scope of the reward function. 

Participants' desire for additional information isn't a shortcoming of our method as participants in the Local Experiential Explanation group did well on the predictive task; it shows areas for future work addressing the multi-dimensional nature of users' needs in complex environments. Since influence predictors tracked the sources of reward for the agent, local Experiential Explanations extracted and provided participants with all the information possible from the model. The information we provided about the circumstances related to the agent's actions still significantly helped participants predict the circumstances leading to the agent's actions. Conseuqently, the request of some participants can help identify where new explanation methods can be developed to start meeting users' other needs. Looking to the codes around present and missing features helps highlight some of these directions, including users' desires around the \textit{Game} (i.e., \textcode{blue}{Goals, Rules, World}) and their desires around the \textit{Agent} (i.e, \textcode{blue}{Actions, Strategies, and Goals}), which were echoed across all experimental groups. 

\myparagraph{Examining How Participants Attempt to Fill the Gaps to Complete their Understanding of the Agent.}
Because of the extreme disguise of the tasks, participants expressed an unanticipated degree of creativity and curiosity to explain the actions of the agent. The environment in which we placed the participants was intentionally opaque to ensure that they relied on the information provided in the explanations, rather than relying on their knowledge from other games or experiences. Still, participants used their imaginations to fill in the gaps around what was not provided in the explanations and reconcile the agent's model with their knowledge. For example, one pariticipant hypothesized that the grey ducks were actually ``evil robot ducks'' (L12). Another participant created a full narrative around the explanation about how Jade is trying to escape a cave (N16). There was also a plethora of hypotheses around healing apples (L25), or Jade's desire for a new pet lizard or crab (N25, L26, A27, L31). We set up Jade as an agent in an environment that would not make intuitive sense, giving participants the job of predicting her task with an explanation. The participants filled out her story around the explanations they were given, adding motive and context where we did not.

In some contexts, creativity is not a desired feature. For example, it is (probably) not good for doctors (or others) to creatively invent reasons around AI recommendations for medical environments. Similarly, users should not be creatively attributing reasons their bank loan was rejected. However, in storytelling and games, creativity and curiosity are critical features of a positive and engaging experience. This opens up the direction of future work for these environments to understand how explanations can be a feature that encourages creativity and curiosity while maintaining high task performance and understanding of the agent and promoting an overall positive experience. We encourage future work to consider exploring explanation methods as places to which people can tether their imaginations, building on explanations to construct fun, interesting storytelling experiences.

\section{Limitations and Future Work}

Our experiments focus on the content of the explanation rather than optimizing the language delivery or comparing different styles of language explanations. 
Templates were sufficient to show a significant affect on human participants regardless of the acknowledged limitations of templates such as rigidity and formulaicity of text.
As such, we evaluated against baselines with other presentational modalities, such as visual or numeric modalities, instead of language variation baselines.
As more verbal and fluent explanation methods are developed, more research should consider how different presentational aspects of language generation can impact users' performance and preference for explanation. 
Preliminary investigation by \citeauthor{tambwekar2024towards} \cite{tambwekar2024towards} suggests that surface-form language variation may not have a large impact.

Our approach highlights broader trends in the influences presenting an overview of how they collectively impact the agent's actions over time. While this provides a helpful summary, there is potential to derive richer insights by examining these interactions in greater depth. In certain contexts, the interactions between these influences may provide valuable insights for system users. Nonetheless, it remains unclear whether such detailed explanations are feasible or necessary.

We do not consider the question of {\em when} users should ask for explanations; 
we make the simplifying assumption that an explanation generation system should be able to generate an explanation at any point in response to any ``why not'' question from an operator for any action performed by the agent. 
However, further research is required to understand the cognitive load involved in explanations that are presented in real-time in an interactive context. 
Determining the optimal timing and method for presenting these real-time insights will be crucial to ensure they enhance user understanding without causing undue strain.
For example, explanations given in real-time in an environment like Crafter might be disruptive, whereas an {\em after-action review} \cite{dodge2021after} may allow an operator to revisit a previous point in time after the fact to seek explanations.

Future research in this domain should explore the potential of the multi-dimensional nature of explanations to more effectively reconcile users' mental models of agents by addressing the disconnects in knowledge that might arise in understanding any part of the system. This exploration can include the integration of various sources of explanation, different modalities, and innovative interaction techniques. For example, exploring how increased interactivity and allowing users to ask varying questions at different times would have impacts on the users' performance and preference of explanations \cite{qian2022evaluating}. Explainable RL will be a crucial component as reinforcement learning algorithms begin to drive agents and rise of human-robots collaboration settings. Future research should investigate how this approach could be extended to continuous environments, robotics, or multi-agent systems to enhance its scope and applicability.

\section{Conclusions}

Explaining reinforcement learning to non-AI expert audiences is challenging. Actionable local explanations should reference future anticipated interactions with respect to the environment. Our Experiential Explanations technique attempts to capture the ``experiences'' of the agent in terms of reward influences that can be relayed to the user in the form of explanations that help them construct an appropriate mental model of how the agent interacts with the environment. Experiential Explanations provide context to the RL agent's behaviors without changing the agent architecture while still providing a high degree of faithfulness. This is significant because any core RL algorithm can be used to train the influence predictors, as we demonstrated through our studies using both Deep $Q$-Networks and PPO. Quantitative and qualitative studies with non-AI expert study participants show that Experiential Explanations help users better understand the agent and predict its behaviors in partially-observable long-horizon discrete planning problems with sparse rewards. They also show that Experiential Explanations are preferred and found more useful than the compared agent explanation baselines. The experimental results suggest that Experiential Explanations may provide for more actionable ways for non-AI expert users to understand the behavior of reinforcement learning agents. 

\bibliography{sn-bibliography}
\appendix

\appendix
\section{Implementation details of the Minigrid Agent}
\label{appendix:agent}
\begin{figure}[!htbp]
    \centering
    \includegraphics[width=\columnwidth]{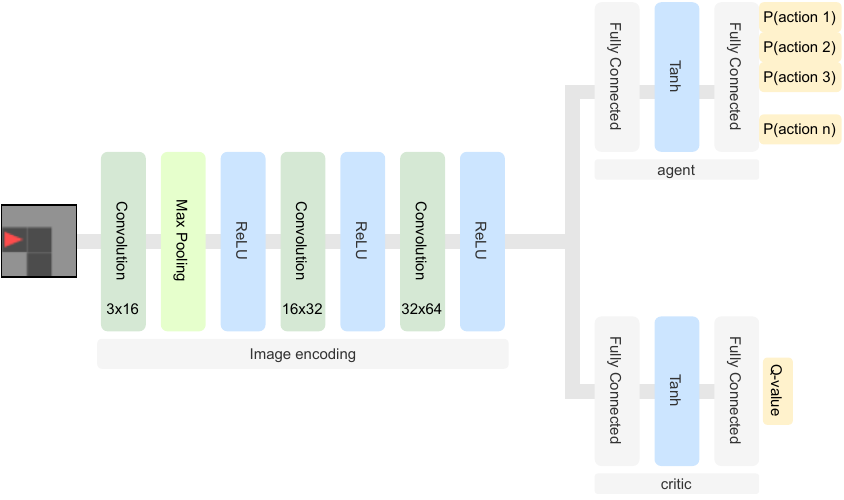}
    \caption{The architecture of the agent A2C model}
    \label{fig:agent-archi}
\end{figure}

Figure \ref{fig:agent-archi} shows the architecture of our agent. The state observation is encoded using a convolutional neural network (CNN), then passed to the actor-critic network heads composed of two shallow MLPs with one hidden layer each. The Minigrid agent was implemented using the torch-ac implementation presented in the official Minigrid documentation. \footnote{https://github.com/lcswillems/rl-starter-files/}

We train our agents with the following hyperparameters:
\begin{align*}
\text{processes} &= 20 \\
\text{frames} &= 1221120 \\
\text{batch size} &= 256 \\
\text{discount factor} &= 0.99 \\
\text{learning rate} &= 3.5 \times 10^{-4} \\
\textbf{PPO hyperparameters:} \\
\text{epochs} &=10 \\
\text{lambda coefficient in GAE} &= 0.95\\ 
\text{entropy term coefficient} &=0.01\\
\text{value loss term coefficient} &=0.5\\
\text{maximum norm of the gradient} &=0.5\\
\text{optimizer} &= Adam \\
\text{optimizer epsilon} &=1 \times 10^{-8}\\
\text{clipping epsilon for PPO}  &=0.2\\
\end{align*}

\section{Implementation details of the Minigrid Influence Predictors}
\label{appendix:influence_predictor}
\begin{figure}[!htbp]
    \centering
    \includegraphics[width=\columnwidth]{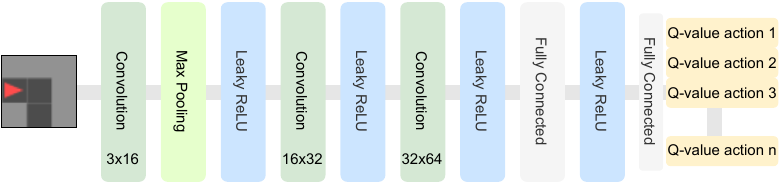}
    \caption{The architecture of the influence predictor network.}
    \label{fig:influenceDQNarchitecture}
\end{figure}
Our DQN influence predictors are implemented in Pytorch \footnote{https://pytorch.org/tutorials/intermediate/reinforcement\_q\_learning.html} and all the code is available in our git repository. Figure \ref{fig:influenceDQNarchitecture} details the architecture of the influence predictor's model. We train the influence predictor with the following hyperparameters:
\begin{align*}
    \text{replay buffer capacity} &= 50000  \\
\text{gamma} &= 0.5 \\
\text{target update} &= 5  \\
\text{bootstrap} &= 100000  \\
\text{batch size} &= 64 \\
\text{learning rate} &= 1 \times 10^{-5} \\
\text{optimizer} &= \text{RMSprop}
\end{align*}
\section{Implementations details of Crafter's Agent and Influence Predictors}
\label{appendix:crafteragent}
As mentioned earlier, both the Agent and the Influence Predictors are implemented using Stable-Baselines3's \cite{stable-baselines3} PPO \cite{schulman2017proximal} implementation. Given the observations are in the form of images, the policy used here ActorCriticCnnPolicy (also from the Stable-Baselines3 \cite{stable-baselines3}).

They use the same values as hyperparameters. The training was done on a single node of a single a40 processor. The training time can be reduced by parallelizing over multiple nodes/processors. The hyperparameters were obtained from \cite{stanić2022learning}.

Hyperparameters for the implementations: 
\begin{align*}
\text{nodes} &= 1 \\
\text{timesteps} &= 8994816 \\
\textbf{PPO hyperparameters:} \\
\text{learning rate} &= 3 \times 10^{-4} \\
\text{Number of steps per env per update} &= 4096 \\
\text{gamma} &= 0.95 \\
\text{Number of epochs} &= 4 \\
\text{batch size} &= 128 \\
\text{lambda coefficient in GAE} &= 0.65\\
\text{entropy term coefficient} &= 0.0\\
\text{value loss term coefficient} &= 0.5\\
\text{maximum norm of the gradient} &= 0.5\\
\text{optimizer} &= Adam \\
\text{clipping parameter}  &= 0.2\\
\end{align*}

\section{Experiment 1: Evaluation Survey}
\label{appendix:eval-survey1}
\begin{figure}[!htbp]
    \centering
    \includegraphics[width= \columnwidth]{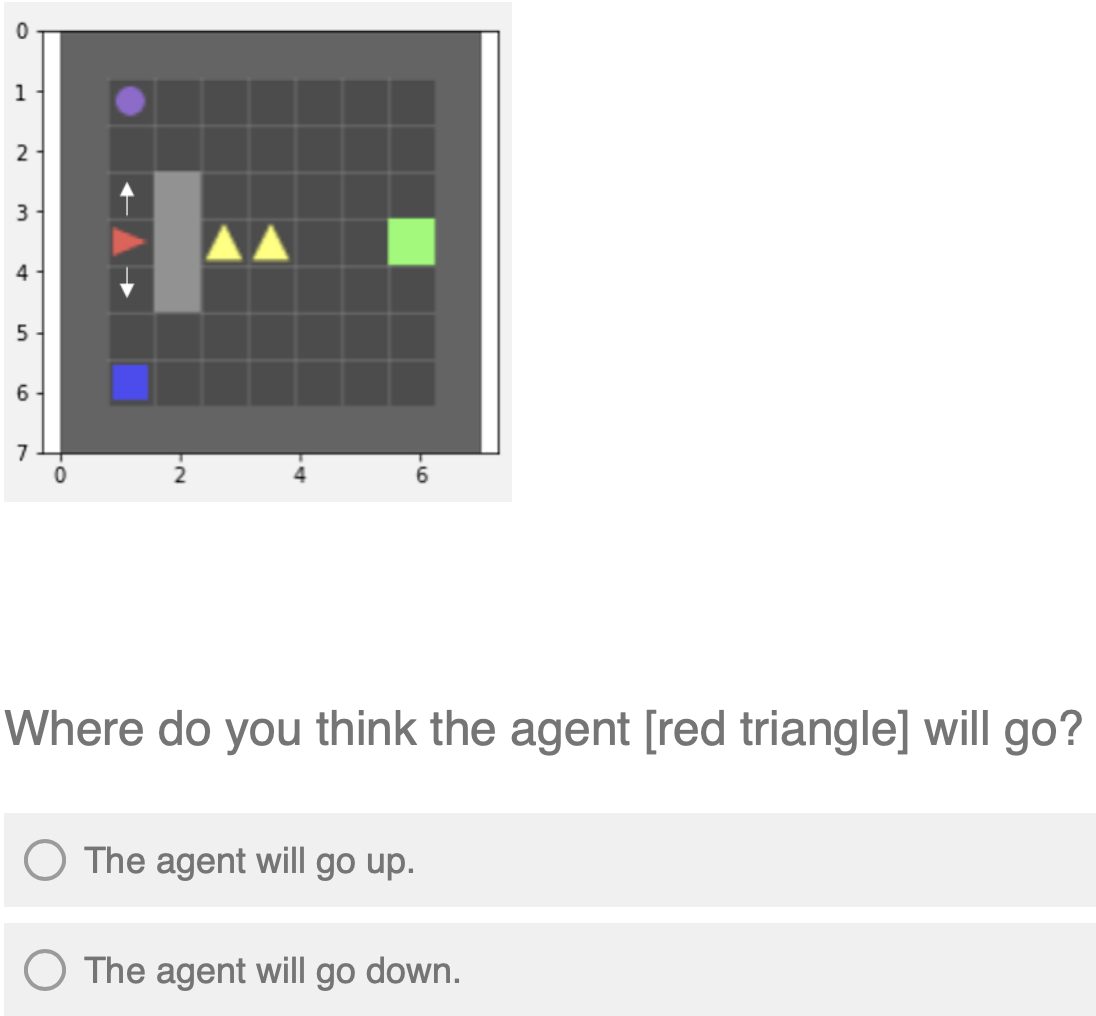}
    \caption{This figure shows an example of the prediction task. The participants saw the map and predicted which direction they thought the agent preferred.}
    \label{fig:prediction-question}
\end{figure}

The Human Evaluation Study in Minigrid is made up of three parts. The first part is a prediction task, as shown in the example in Figure \ref{fig:prediction-question}. We presented three episodes to each participant, as shown in Figure~\ref{fig:cases}. During each episode, the agent---the red triangle---must navigate to the green square. Some colored shapes are associated with negative rewards, but the participant is not told which ones. 
After each episode, the participant is asked to predict whether the agent will go up or down.
The participant has no information at the beginning of the first episode to make an informed choice, and the participant choice is random---we do not evaluate on this choice. 
Further, upon the participant's choice, we assign which object is dangerous such that the participant is always wrong. 
This established the conditions for which the participant must now learn to correct their understanding of the agent's interactions with the environment.
We use that assignment of dangerous object consistently throughout the second and third episodes.
The participant tries to predict the agent's choice after the second and third episodes (without the object assignment switching); we expect to see the participants in the experimental condition to have higher accuracy rates than in the baseline conditions.

\begin{figure}[!htbp]
    \centering
    \includegraphics[width= \columnwidth]{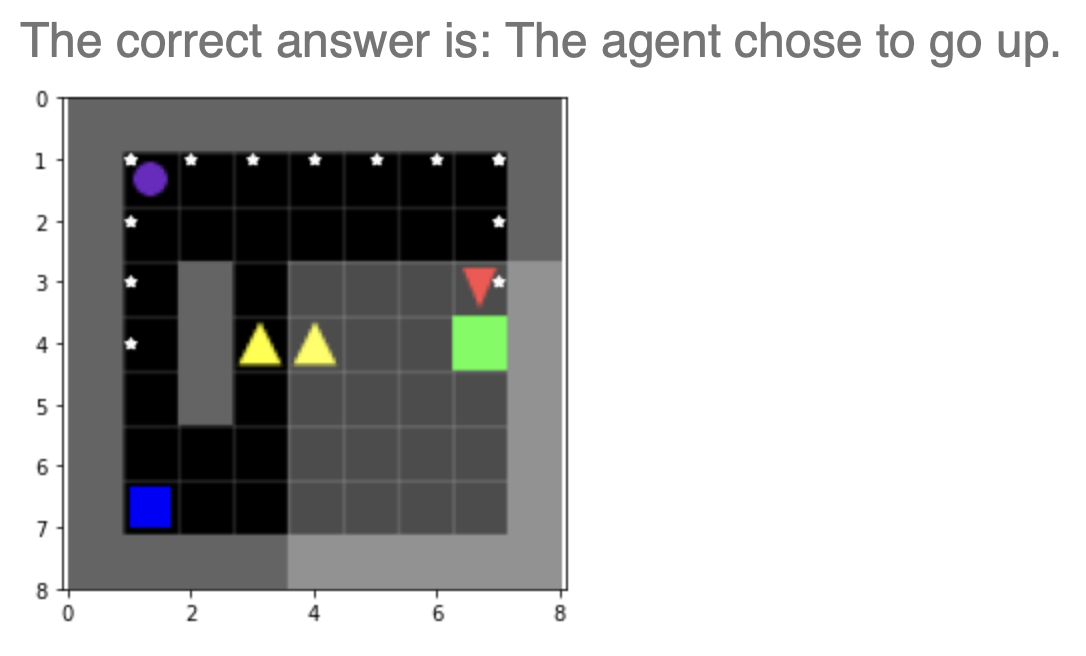}
    \caption{After predicting the agent's following action, the participants sees the correct actionseeis is the only information provided to the No Explanation group.}
    \label{fig:correct-answer}
\end{figure}

\begin{figure}[!htbp]
    \centering
    \includegraphics[width= \columnwidth]{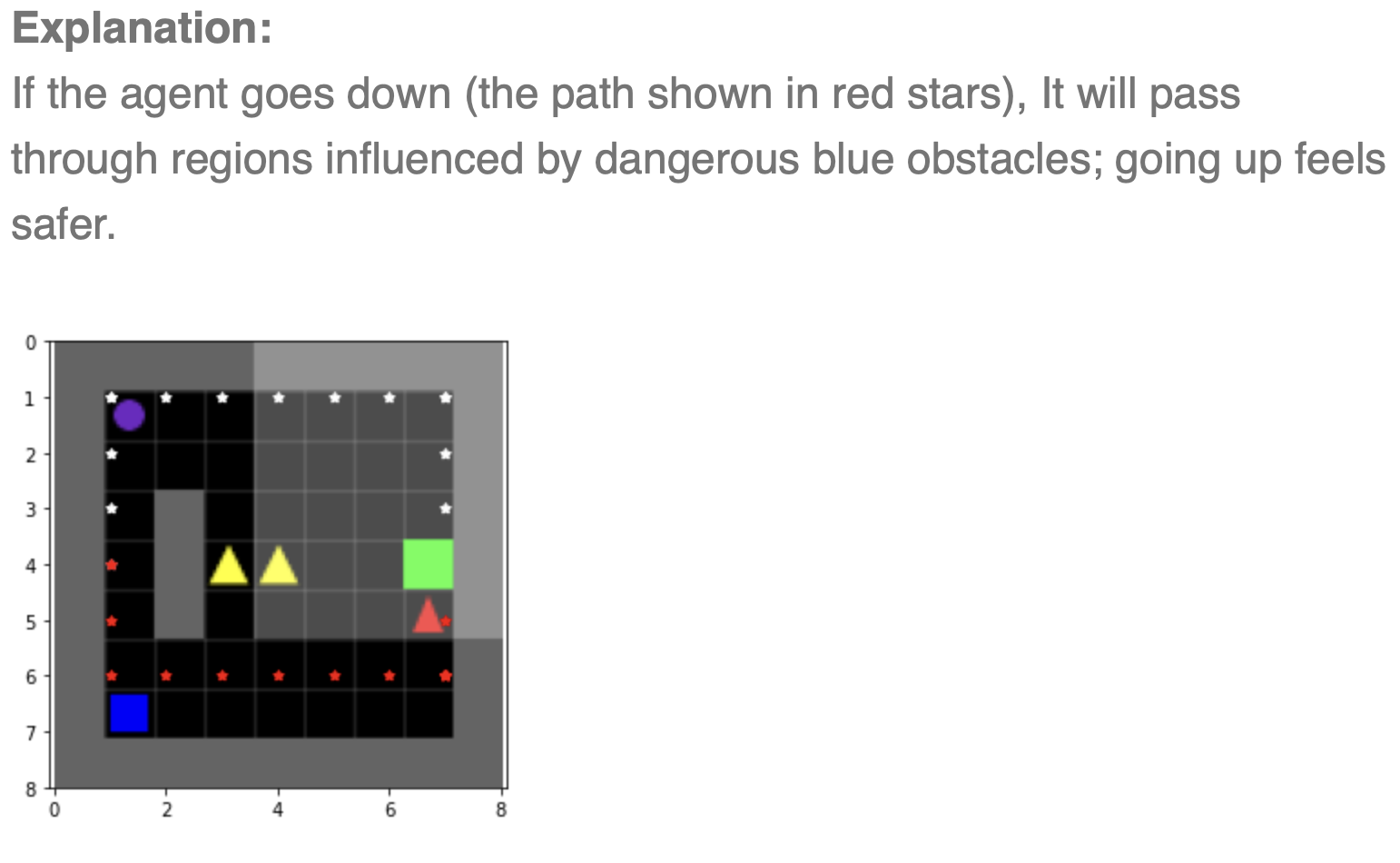}
    \caption{This Experiential Explanation highlights the dangerous objects and compares the agent's actual and suggested path.}
    \label{fig:ee}
\end{figure}

\begin{figure}[!htbp]
    \centering
    \includegraphics[width= \columnwidth]{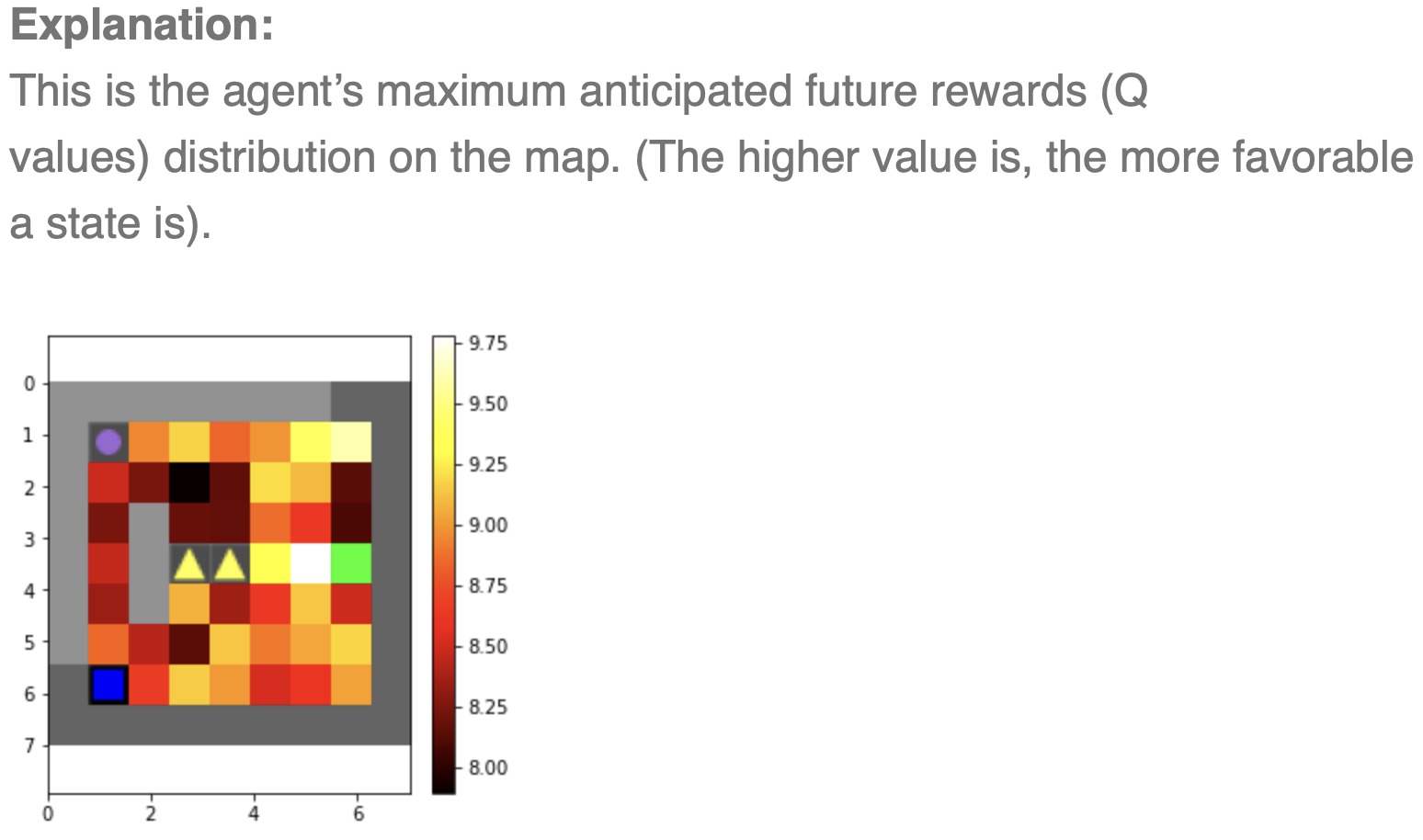}
    \caption{This heat map explanation shows the maximum anticipated future reward for each state in the environment.}
    \label{fig:he}
\end{figure}

\begin{figure}[!htbp]
    \centering
    \includegraphics[width= \columnwidth]{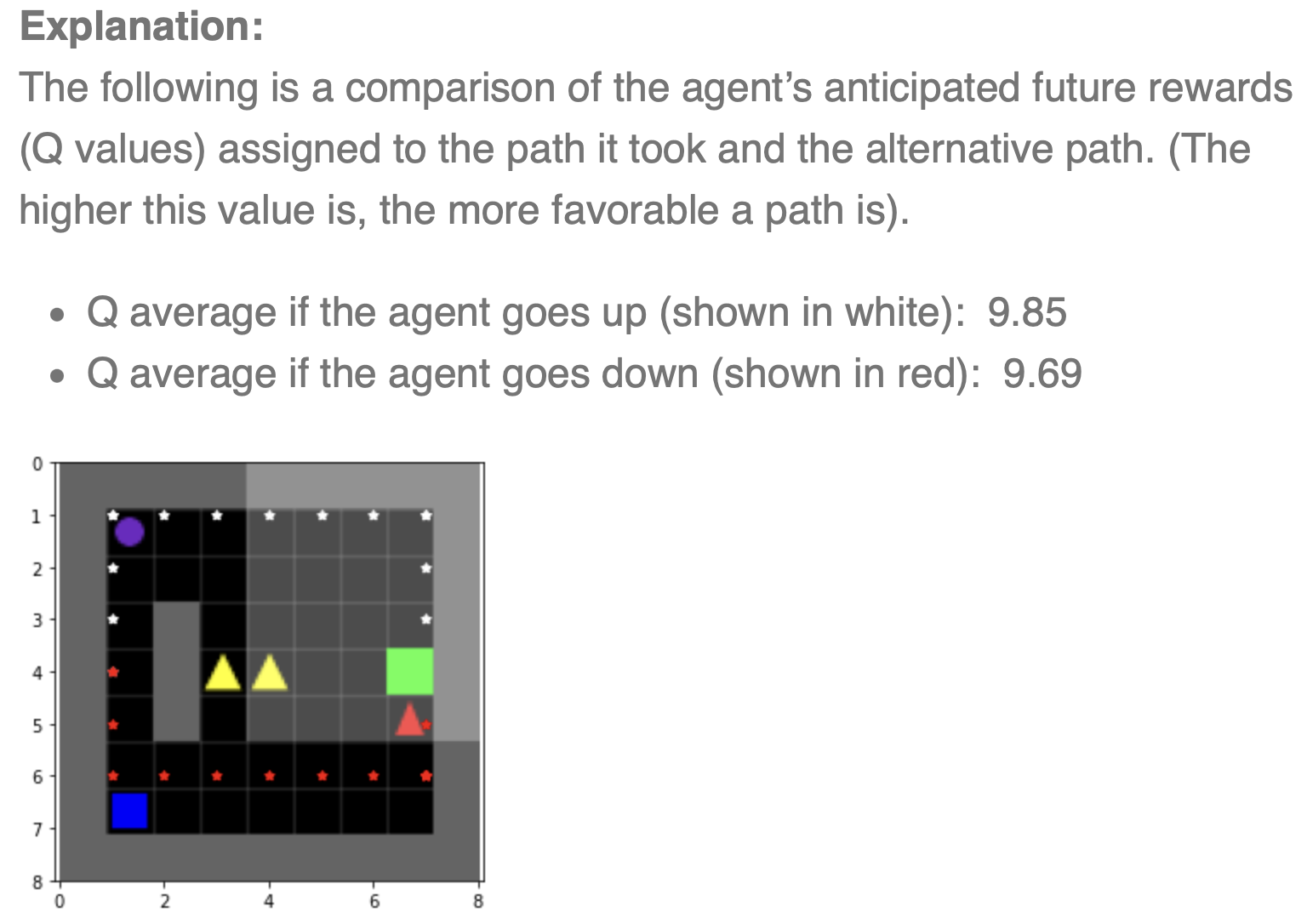}
    \caption{This Q-value explanation compares the actual and suggested paths regarding their average anticipated future reward.}
    \label{fig:qe}
\end{figure}

After each episode, participants fill out a brief text entry question explaining the rationale behind their choice. Then after the participants submit their predictions and reasoning, they see the correct answer as shown in Figure \ref{fig:correct-answer}. 
Afterward, they see one of the three types of explanations, Experiential Explanation (Figure \ref{fig:ee}), Heatmap Explanation (Figure \ref{fig:he}), Q-value explanation (Figure \ref{fig:qe}), or they see no explanation at all. Then they repeat this process two times. 

\begin{figure}[!htbp]
    \centering
    \includegraphics[width= \columnwidth]{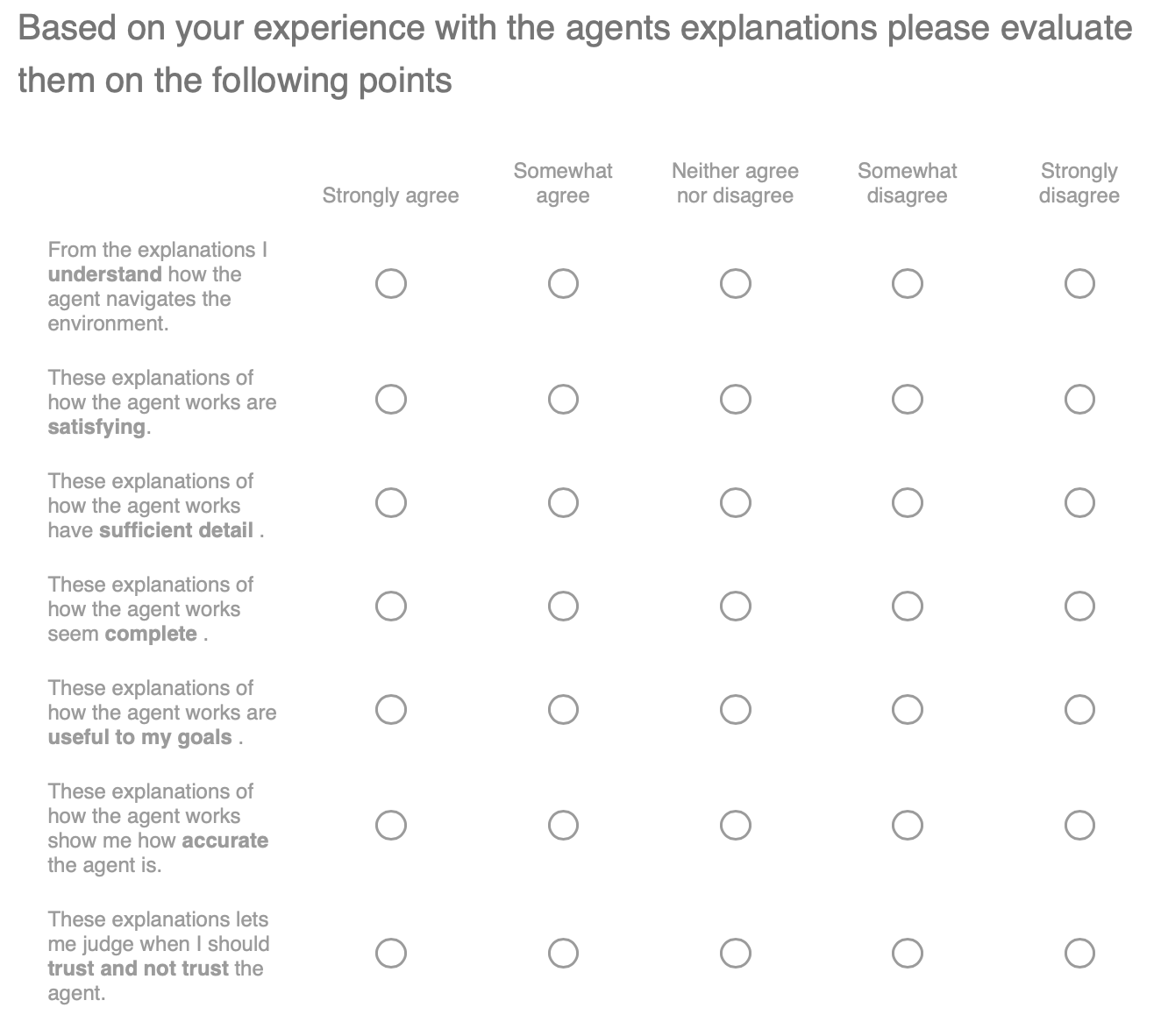}
    \caption{After the experiments, the participants completed the shown satisfaction survey evaluating the explanation type they interacted with on a seven-point scale, measuring helpfulness, satisfaction, understandability, level of detail, completeness, accuracy, and trust. }
    \label{fig:ss}
\end{figure}

After the participants finish the prediction task, they proceed to the second part of the study, a satisfaction survey, consisting of five-point Likert scale questions asking the user to rate understandability, satisfaction, amount of details, completeness, usefulness, precision, and trust for the explanations they saw. It also had an open-ended response question asking how the explanation helped or did not help them understand the agent. This survey is shown in figure \ref{fig:ss}.


\section{Experiment 1: Satisfaction Survey Analysis}
\begin{figure*}[!tbp]
    \centering
    \includegraphics[width= \textwidth]{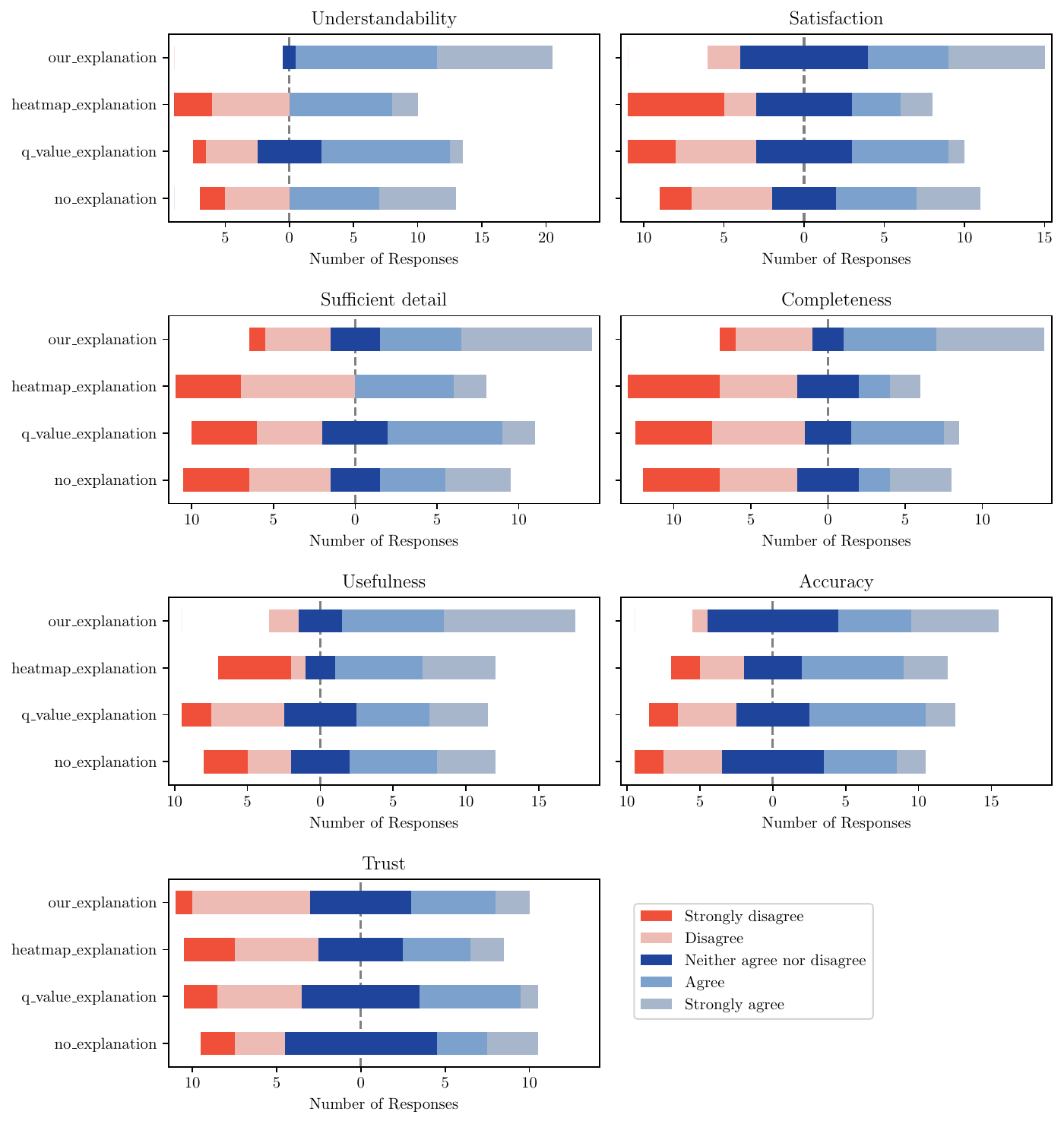}
    \caption{Experiment 1 satisfaction survey results for each explanation quality dimensions grouped by explanation type}
    \label{fig:survey}
\end{figure*}

Figure \ref{fig:survey} shows the results of the satisfaction survey questions shown in Figure \ref{fig:ss}. The discussion and analysis of the Figure \ref{fig:survey} results are in the main paper.

\label{appendix:quant-analysis2}

\section{Crafter Re-skin mappings}
\begin{table}[t]
    \centering
    \footnotesize
    \begin{tabular}{llc c}
        \toprule
            Actual Asset & Re-skinned Asset \\
        \midrule
            Materials and Objects\\
        \midrule
            Player (Jade) & \emph{No change}\\
            Tree & Rock\\
            Grass & \emph{No change} \\
            Sand & \emph{No change}\\
            Water & \emph{No change}\\
            Stone & Scissors\\
            Iron & Paper\\
            Coal & Lizard\\
            Diamond & Electric torch\\
            Zombie & Duck\\
            Table & Parrot\\
            Furnace & Crab\\
            Bridge & \emph{No change}\\
            Cow & Apple\\
        \midrule
            Inventory Items\\
        \midrule
            Wood & Rock slabs\\
            Wooden pickaxe & Cutter\\
            Wooden sword & Throwing knives\\
            Iron pickaxe & Coaster\\
            Iron sword & Fan\\
            stone pickaxe & Plier\\
            stone sword & Hammer\\
        \midrule
            Parameters\\
        \midrule
            Health & \emph{No change}\\
            Food & \emph{No change}\\
            Drink & \emph{No change}\\
            Energy & \emph{No change}\\
        \bottomrule
    \end{tabular}
    \caption{Original Crafter assets vs their Rock-Paper-Scissors (RPS) asset collection counterparts}
    \label{tab:crafter-asset-reskins}
\end{table}
Table \ref{tab:crafter-asset-reskins} lists a complete mapping of assets re-skinned assets

\section{Experiment 2: Evaluation Survey}
\label{appendix:eval-survey2}

\begin{figure}
    \centering
    \includegraphics{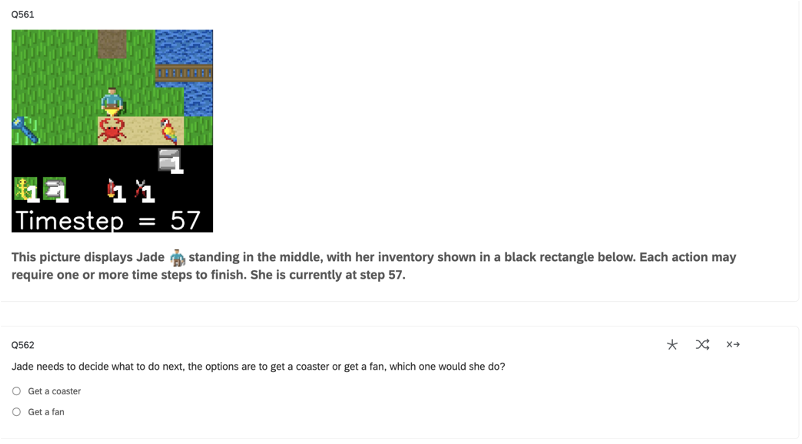}
    \caption{The first decision point the participants encounter during experiment 2.}
    \label{fig:exp2_predict1}
\end{figure}

\begin{figure}
    \centering
    \includegraphics{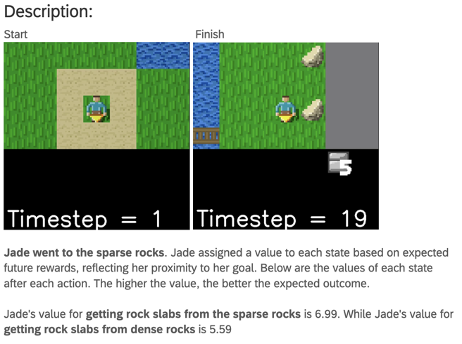}
    \caption{An example of the agent value explanation baseline.}
    \label{fig:exp2_agent_value_explanations}
\end{figure}

\begin{figure}
    \centering
    \includegraphics{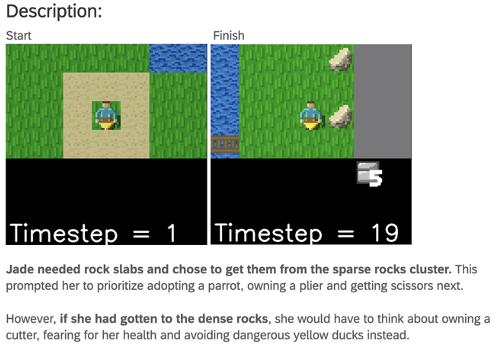}
    \caption{An example of the local experiential explanation.}
    \label{fig:exp2_local_explanations}
\end{figure}

The Human Evaluation Study in Crafter has three tasks, a prediction task and an ordering task, and a satisfaction survey. We picked four random decision points from the agent's trajectory, each paired with a counterfactual trajectory that the agent can take by performing a counterfactual action at this decision point. We presented each participant with one decision point and asked them to predict the agent's action as shown in Figure \ref{fig:exp2_predict1}. After this first interaction, they are shown their assigned explanation. The explanations are one of the three baselines specified in Section \ref{sec:exp2_baselines}. An agent value explanation (Figure \ref{fig:exp2_agent_value_explanations}), the local experiential explanation (Figure \ref{fig:exp2_local_explanations}), or they do not see an explanation.

\begin{figure}
    \centering
    \includegraphics{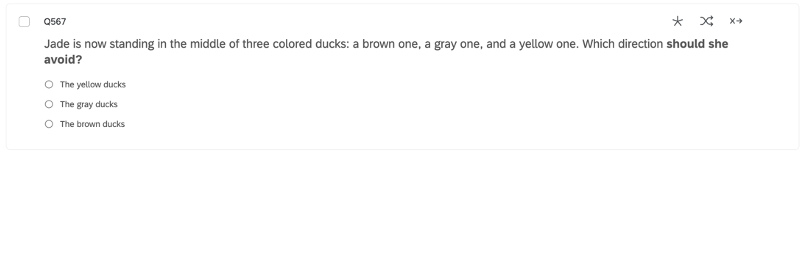}
    \caption{The second decision point the participants encounter during experiment 2. Which is used for the prediction task and followed by the reasoning question.}
    \label{fig:exp2_predict2}
\end{figure}

\begin{figure}
    \centering
    \includegraphics{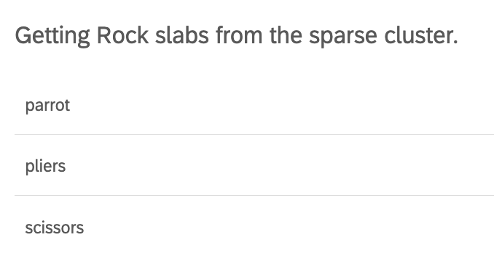}
    \caption{The ordering task which asks the participants to order the influences by their importance to the agent at the decision point.}
    \label{fig:exp2_order}
\end{figure}

Then after seeing the explanation, we asked them to predict and reason about the agent's action at a future decision point displayed in Figure \ref{fig:exp2_predict2}. Lastly, the participants engage in an additional task in order of the influences by their importance to the agent when it considers performing the counterfactual action or the actual action as shown in Figure \ref{fig:exp2_order}. 
Both tasks are followed by a text entry question asking them to reason about their predictions.

After participants complete both tasks, they proceed to the last part
of the study, a satisfaction survey identical to the one we used in experiment 1 as shown in Figure \ref{fig:ss}. consisting of five-point Likert scale questions asking
the user to rate understandability, satisfaction, amount of details, completeness, usefulness, precision, and trust for the explanations they saw. It also had an open-ended
response question asking how the explanation helped or did not help them understand
the agent. The results of the survey for Experiment 2 is shown in Figure \ref{fig:survey}.

\begin{figure}
    \centering
    \includegraphics[width=\columnwidth]{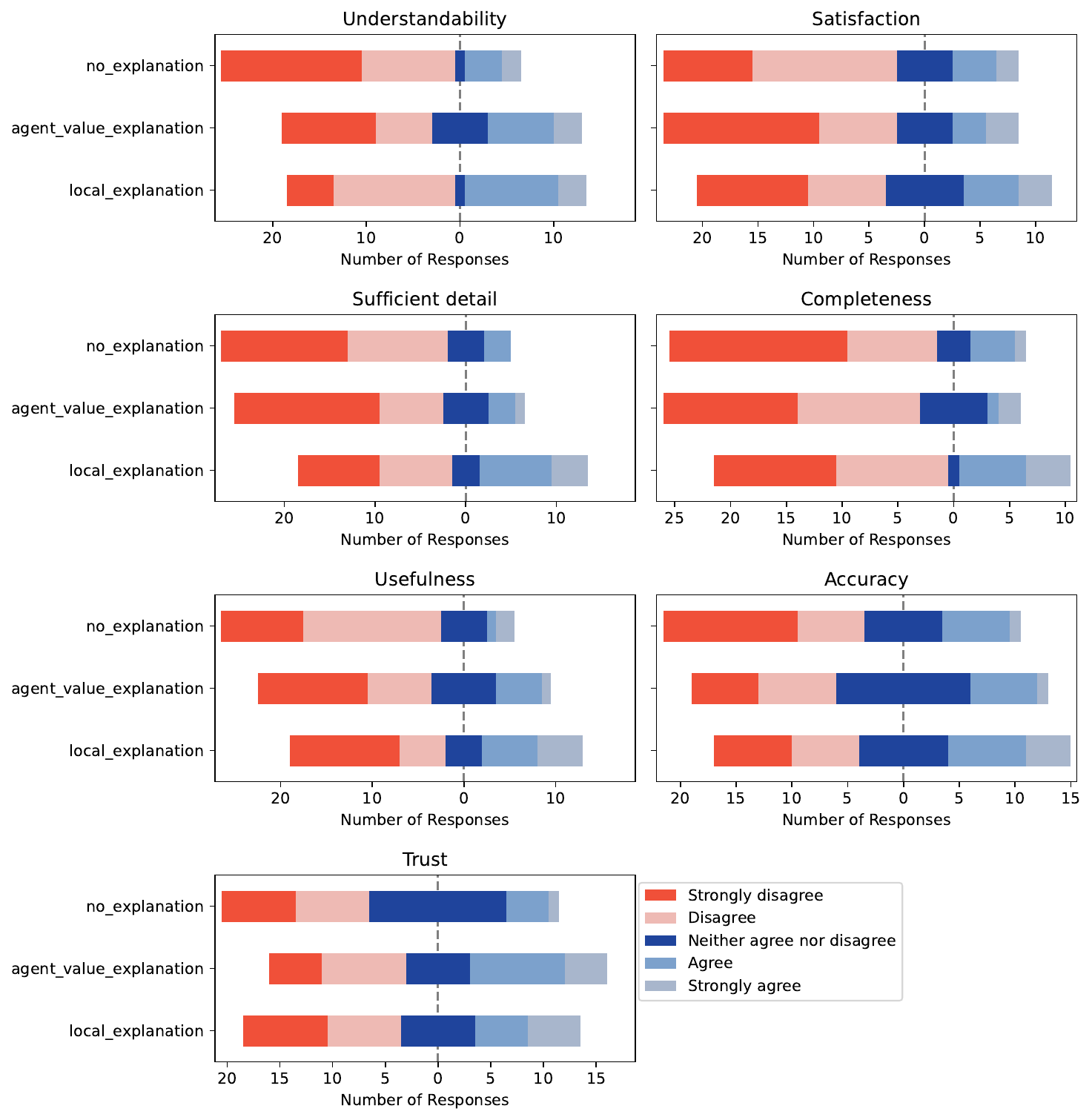}
    \caption{Experiment 2 satisfaction survey results for each explanation quality dimensions grouped
by explanation type}
    \label{fig:ex2-survey}
\end{figure}

\end{document}